\documentclass{article}

\usepackage[main, preprint]{neurips_2025}
\usepackage[T1]{fontenc}    % use 8-bit T1 fonts
\usepackage{hyperref}       % hyperlinks
\usepackage{url}            % simple URL typesetting
\usepackage{booktabs}       % professional-quality tables
\usepackage{amsfonts}       % blackboard math symbols
\usepackage{nicefrac}       % compact symbols for 1/2, etc.
\usepackage{microtype}      % microtypography
\usepackage{xcolor}         % colors
\usepackage{lipsum}         % Just for generating dummy text, can be removed
\usepackage{amsmath}
\usepackage{graphicx}

\usepackage{multirow}
\usepackage{booktabs}
\usepackage{amssymb}
\usepackage{enumitem}
\usepackage{marvosym}
\usepackage[table]{xcolor}
\usepackage{caption}
\usepackage{subcaption}
\usepackage{makecell}
\usepackage{tabularx}
\usepackage{pifont}
\usepackage{fontawesome5}
\usepackage{float}

\newcommand{\zhangsqzhangsh}[1]{\textcolor{black}{#1}} % 已确认

\title{\textsc{EdgeRazor}: A Lightweight Framework for Large Language Models via Mixed-Precision Quantization-Aware Distillation}

\author{%
  Shu-Hao Zhang\textsuperscript{1,2} \quad
  Le-Tong Huang\textsuperscript{1,2} \quad
  Xiang-Sheng Deng\textsuperscript{1,2} \quad
  \textbf{Xin-Yi Zou\textsuperscript{4}} \\[0.2em]
  \textbf{Chen Wu\textsuperscript{4}} \quad
  \textbf{Nan Li\textsuperscript{4}} \quad
  \textbf{Shao-Qun Zhang\textsuperscript{1,2,\Letter}} \quad
  \textbf{Zhi-Hua Zhou\textsuperscript{1,3}}  \\[0.3em]
  \small \textbf{\textsuperscript{1}} National Key Laboratory for Novel Software Technology, Nanjing University, Nanjing 210063, China \\
  \small \textbf{\textsuperscript{2}} School of Intelligent Science and Technology, Nanjing University, Suzhou 215163, China \\
  \small \textbf{\textsuperscript{3}}School of Artificial Intelligence, Nanjing University, Nanjing 210063, China \\
  \small \textbf{\textsuperscript{4}} Microsoft AI, Beijing 100080, China \\
  \texttt{ \{zhangsh,zhangsq\}@lamda.nju.edu.cn }
}
\begin{document}

\maketitle

\vspace{-15pt}
\begin{abstract}
Quantization has emerged as a mainstream approach for deploying Large Language Models (LLMs) on resource-constrained devices, yet compressing precision below 4-bit typically causes severe performance degradation or prohibitive retraining costs. In this paper, we propose \textsc{EdgeRazor}, a lightweight framework for LLMs via Mixed-Precision Quantization-Aware Distillation. It contains three modules: Structural Quantization with Mixed Precision for fine-grained control of bit-widths, Layer-Adaptive Feature Distillation that dynamically selects the most informative features for alignment, and Entropy-Aware KL Divergence for forward-reverse balance on both human-annotated and distilled datasets. Evaluations conducted on MobileLLM and Qwen families show that under weight-activation quantization, the 1.88-bit Qwen3-0.6B-\textsc{EdgeRazor} outperforms the state-of-the-art 2-bit baselines by 11.27 and surpasses the strongest 3-bit baselines by 4.38, while the quantized MobileLLM-350M-\textsc{EdgeRazor} requires a training budget 4--10$\times$ lower than the leading quantization-aware training method. In terms of efficiency, \textsc{EdgeRazor} achieves higher compression ratios at all bit-widths, and the 1.58-bit Qwen3-0.6B-\textsc{EdgeRazor} reduces storage from 1.11\,GB to 0.19\,GB while accelerating decoding by 15.16$\times$ over the 16-bit baseline. These results empirically validate the effectiveness and efficiency of \textsc{EdgeRazor}. \href{https://github.com/zhangsq-nju/EdgeRazor}{\faGithub~{Code}}
\href{https://huggingface.co/collections/zhangsq-nju/edgerazor-nbit}{\raisebox{-1pt}{\includegraphics[height=1em]{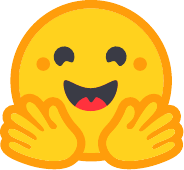}}~{Models}}
\end{abstract}
\begin{figure*}[!ht]
    \centering
    \vspace{-10pt}
    \includegraphics[width=0.92\textwidth]{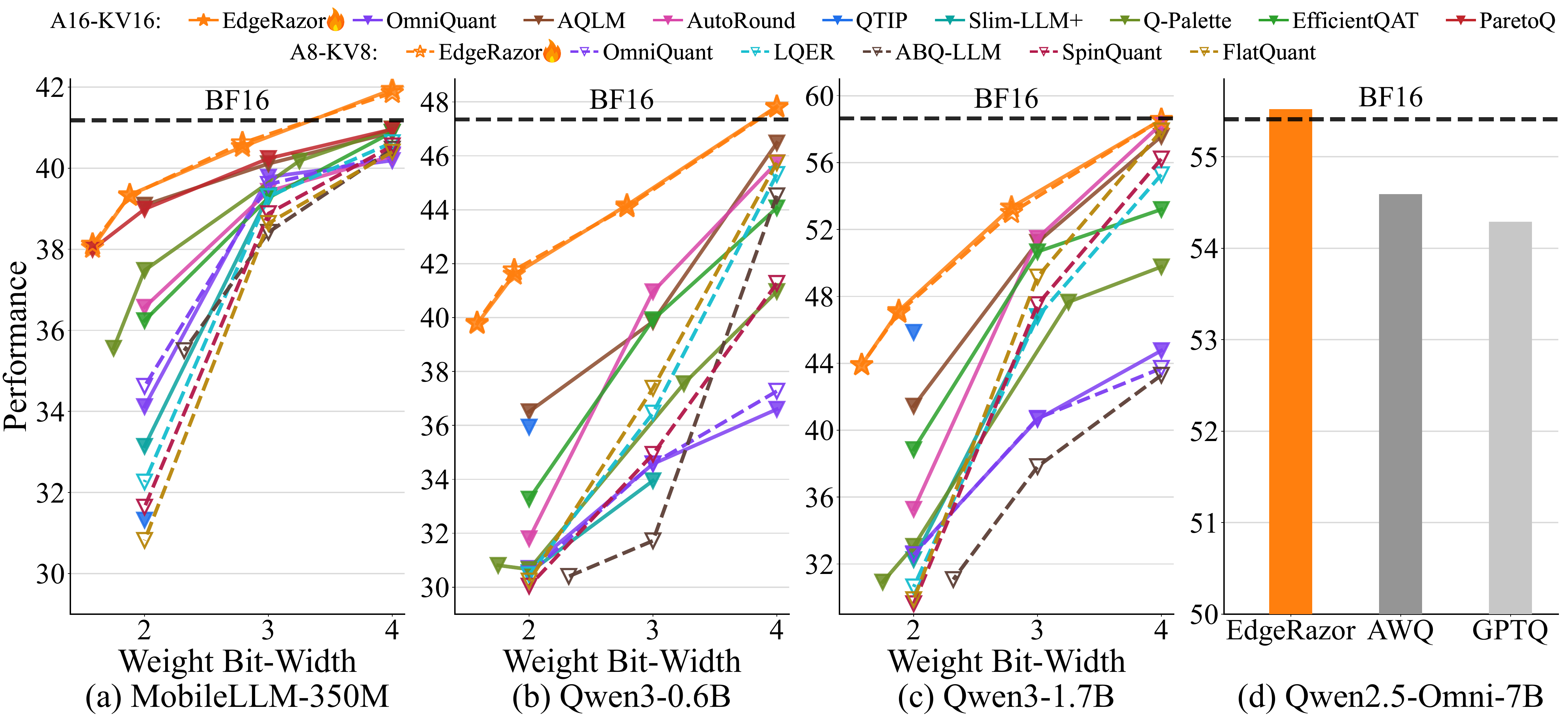}
    \caption{Performance comparison of \textsc{EdgeRazor} and strong baselines.}
    \label{fig:perf}
\end{figure*}

\clearpage
\newpage

%%%%%%%%%%%%%%%%%%%%%%%%%%%%%%%%%%%%%%%%%%%%%%%%%%%%%%%%%%%%

\section{Introduction}

Large language models (LLMs) have attracted widespread attention across domains, driven by scaling laws indicating that performance improves predictably with increases in model size, dataset size, and training budget.
There is an increasing demand to deploy lightweight LLMs on resource-constrained devices, where limited storage, memory, and computational capacity impose constraints that full-precision models struggle to meet~\citep{zheng2025edge}. 
Quantization is an effective compression technique that maps full-precision weights and activations to low-bit representations~\citep{zhu2024survey}. An effective method should satisfy several prerequisites, including preserving performance, offering flexible bit-widths for deployability on resource-constrained hardware, and maintaining manageable training overhead~\citep{tan2024mobilequant}.

% PTQ QAT QAD: how + limit
Existing LLM quantization methods can be broadly categorized into three paradigms: Post-Training Quantization (PTQ), Quantization-Aware Training (QAT), and Quantization-Aware Distillation (QAD).
PTQ calibrates quantized parameters on a small dataset without retraining~\citep{frantar2023gptq,lin2024awq}, while incurring significant performance degradation at lower bit-widths~\citep{dettmers2023case}.
QAT learns low-bit parameters using surrogate gradients~\citep{bengio2013ste} via dataset-driven gradient updates to fit downstream tasks and preserve performance below 4-bit. Nevertheless, QAT incurs substantial training costs~\citep{liu2025paretoq,wang2025bitnet}.
QAD integrates QAT with knowledge distillation~\citep{hinton2015:KD,zhou2004:KD} from a full-precision teacher to reduce the training cost~\citep{liu2023llmqat}. However, QAD methods typically rely on heuristic approaches to pre-specify teacher layers for supervision~\citep{xu2024onebit}, and restrict the switching criterion between forward and reverse KLD exclusively to teacher-distilled data~\citep{du2024bitdistiller}, which precludes flexible data recipes that combine human-annotated and externally distilled corpora~\citep{wu2025rethinking}.
% 混合比特的需求
Additionally, uniform and discrete bit-widths ignore varying parameter sensitivities~\citep{lin2024awq} and restrict diverse deployment budgets~\citep{lee2025qpalette}. While mixed-precision utilizes parameter importance for allocation~\citep{huang2025slimllm}, it is confined to PTQ methods. Since continuous updates cause severe salience drift during training, a more drift-robust mixed-precision strategy is preferred in QAT and QAD methods.

\begin{figure*}[!t]
    \centering
    \includegraphics[width=1\textwidth]{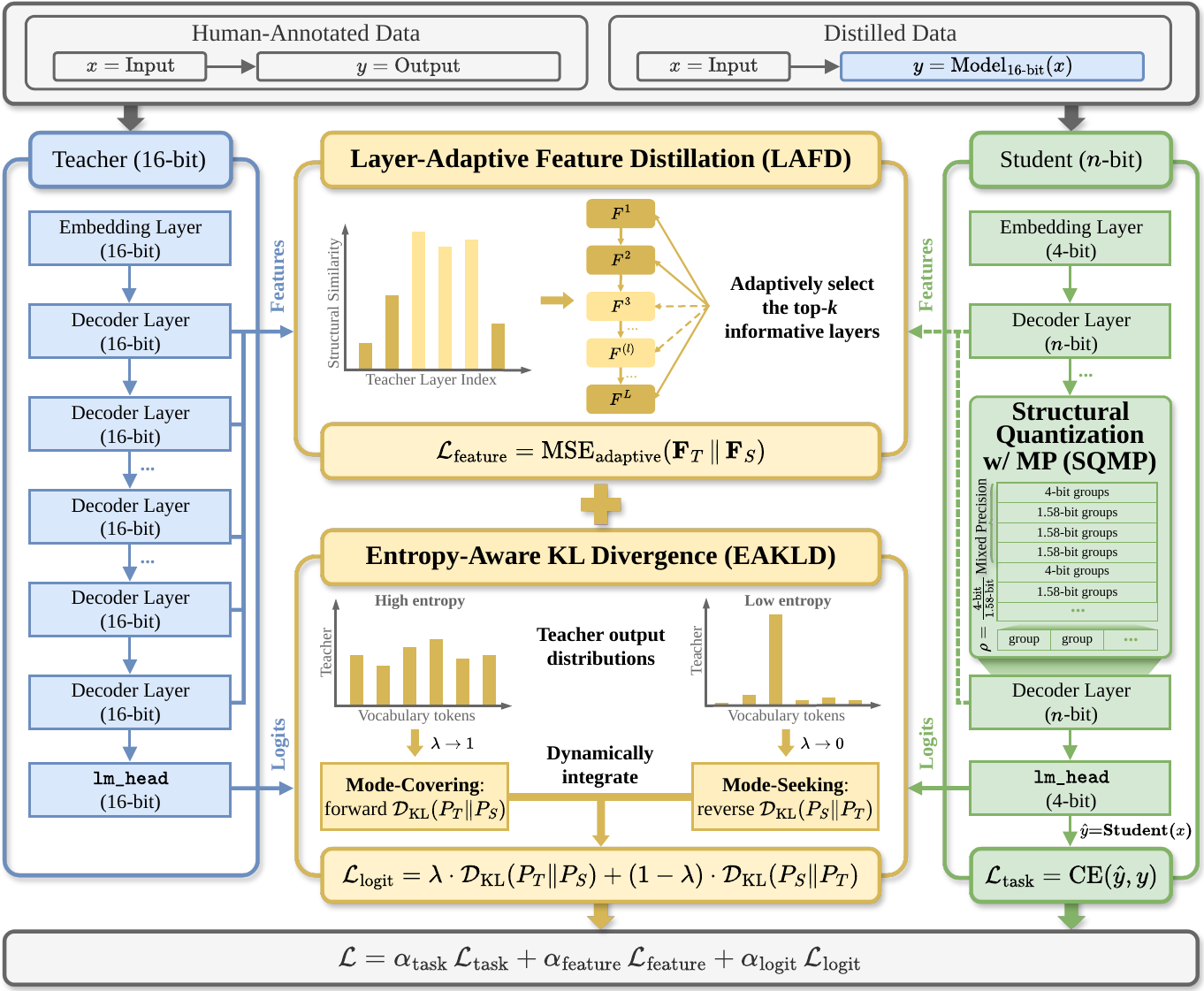}
    \caption{Workflow of the \textsc{EdgeRazor} framework.}
    \label{fig:framework}
\end{figure*}

In this paper, we propose \textsc{EdgeRazor}, a lightweight framework for compressing LLMs with flexible bit-widths via Mixed-Precision Quantization-Aware Distillation (MPQAD). Figure~\ref{fig:framework} illustrates the workflow. \textsc{EdgeRazor} comprises three configurable modules: Structural Quantization with Mixed Precision (SQMP) for fine-grained control over the matrix-wise average bit-width, Layer-Adaptive Feature Distillation (LAFD) that adaptively selects informative teacher layers for feature-level supervision, and Entropy-Aware KL Divergence (EAKLD) that balances forward and reverse KLD by the entropy of the teacher's output distribution, extending logit distillation to both human-annotated and distilled datasets.
% EFFECT then
These modules improve quantized models across base, instruction-tuned, and multimodal LLMs, as summarized in Figure~\ref{fig:perf}.
On Qwen3-0.6B under weight-activation quantization, the 1.88-bit \textsc{EdgeRazor} retains strong performance on downstream tasks, outperforming the state-of-the-art 2-bit PTQ baseline by 11.27 points and surpassing all competing baselines by 4.38 points at 3-bit precision across 14 domain-specific tasks. These performance gains generalize to other architectures, such as MobileLLM-350M, with a training budget 4--10$\times$ lower than that of the leading QAT method.
For deployment, executing the 1.58-bit Qwen3-0.6B-\textsc{EdgeRazor} through \texttt{llama.cpp} on an Apple M4 Pro chip decreases storage from 1.11\,GB to 0.19\,GB, while achieving a 15.16$\times$ decoding speedup over the 16-bit baseline.

The rest of this paper is organized as follows. Section~\ref{sec:related_works} introduces related works. Section~\ref{sec:method} proposes the \textsc{EdgeRazor} framework. Section~\ref{sec:experiments} conducts experiments. Section~\ref{sec:conclusion} concludes this work.

%%%%%%%%%%%%%%%%%%%%%%%%%%%%%%%%%%%%%%%%%%%%%%%%%%%%%%%%%%%%

%%%%%%%%%%%%%%%%%%%%%%%%%%%%%%%%%%%%%%%%%%%%%%%%%%%%%%%%%%%%

% ---------------------------------------------------------

\section{Related works}
\label{sec:related_works}

% \paragraph{Post-Training Quantization.}
PTQ compresses LLMs by calibrating quantized parameters on a small dataset without retraining.
To maintain performance, existing methods employ local error compensation, such as weight adjustments through inverse Hessian approximation~\citep{frantar2023gptq}, activation-aware scaling and outlier smoothing~\citep{lin2024awq,xiao2023smoothquant}, and vector quantization space partitioning~\citep{tseng2024quipsharp}, preserving near-lossless performance at 4-bit and above.
Furthermore, mixed-precision PTQ attempts to optimize structural capacity by heuristically allocating heterogeneous bit-widths across distinct layers or groups~\citep{guan2024aptq,huang2025slimllm,lee2025qpalette}, thereby achieving better accuracy-efficiency trade-offs.
Since calibration-driven strategies lack end-to-end gradient supervision, PTQ often suffers substantial performance degradation in sub-4-bit settings, thereby limiting its viability for resource-constrained deployment~\citep{dettmers2023case}.

% \paragraph{Quantization-Aware Training.}
QAT uses supervised gradient updates with surrogate gradients to optimize low-bit models for target tasks~\citep{bengio2013ste}.
Existing methods typically adopt two paradigms: training natively quantized LLMs entirely from scratch, as pioneered by BitNet~\citep{wang2025bitnet}, and fine-tuning from full-precision pre-trained models via block-wise reconstruction and optimized training budgets, exemplified by EfficientQAT~\citep{chen2024efficientqat} and ParetoQ~\citep{liu2025paretoq}, which significantly advance the frontier of LLM compression to 2-bit and lower.
Nevertheless, these methods typically require substantial computational resources and large-scale corpora to converge~\citep{liu2025paretoq,wang2025bitnet}, rendering the paradigm prohibitively expensive for downstream adaptations.

% \paragraph{Quantization-Aware Distillation.}
QAD integrates QAT with knowledge distillation~\citep{hinton2015:KD,zhou2004:KD} from a full-precision teacher to alleviate the prohibitive computational demands of QAT.
Existing works align output logits and intermediate features, thereby enabling 4-bit compression by data-free generation~\citep{liu2023llmqat} and 1-bit structural decomposition~\citep{xu2024onebit}.
Furthermore, recent advancements combine the mode-covering forward KLD (FKLD)~\citep{hinton2015distilling} with the mode-seeking reverse KLD, utilizing metrics such as teacher prediction confidence to improve zero-shot performance at lower bit-widths~\citep{du2024bitdistiller}.
However, existing QAD approaches are limited by heuristic layer-selection strategies that struggle to generalize across architectures and provide little guidance for selection~\citep{wang2020minilm}, as well as by inflexible KLD-switching criteria that rely exclusively on teacher-distilled data, thereby restricting the use of diverse training recipes~\citep{wu2025rethinking}.

% ---------------------------------------------------------

\section{\textsc{EdgeRazor}}
\label{sec:method}

This section presents \textsc{EdgeRazor}. Figure~\ref{fig:framework} provides an overview of the framework. The \textsc{EdgeRazor} framework for compressing LLMs via MPQAD consists of three novel modules: SQMP that organizes per-channel bit-widths into a periodic super-group pattern with an adjustable ratio in Section~\ref{subsec:method_mp}, LAFD that aligns intermediate representations between student and teacher models by dynamically identifying the most informative layers to supervise in Section~\ref{subsec:LAFD}, and EAKLD that relies solely on the teacher's output distribution to integrate forward and reverse KLD in Section~\ref{subsec:EAKLD}. The overall training objective combines the low-bit student's task-specific cross-entropy loss $\smash{\mathcal{L}_{\mathrm{task}}}$ with the feature and logit distillation losses
\begin{equation}
  \label{eq:total_loss}
  \mathcal{L} = \alpha_{\mathrm{task}}\,\mathcal{L}_{\mathrm{task}} + 
  \alpha_{\mathrm{feature}}\,\mathcal{L}_{\mathrm{feature}} + 
  \alpha_{\mathrm{logit}}\,\mathcal{L}_{\mathrm{logit}} \,,
\end{equation}
where $\smash{\mathcal{L}_{\mathrm{feature}}}$ and $\smash{\mathcal{L}_{\mathrm{logit}}}$ correspond to LAFD and EAKLD, respectively, and $\smash{\alpha_{\mathrm{task}}}$, $\smash{\alpha_{\mathrm{logit}}}$, $\smash{\alpha_{\mathrm{feature}}}$ are balancing coefficients.

\subsection{Structural quantization with mixed precision}
\label{subsec:method_mp}

Based on the per-group quantization function $Q_{n\text{-bit}}(\mathbf{X})$ detailed in Appendix~\ref{apx:quant-func} and the QAD paradigm, we propose SQMP to determine the effective bit-width.
To control bit-widths, SQMP assigns a tunable parameter $\rho\in[0,1]$ indicating the proportion of weights assigned to 4-bit.
We organize this structural mixed-precision quantization into a regular repeating super-group allocation: every $\lfloor 1/\rho \rceil$ consecutive output channels form one super-group, wherein one channel is quantized to 4-bit and the remainder to 1.58-bit. The illustration is provided in Appendix~\ref{apx:mpqad-allocation}.
Since every super-group maintains the same configuration, altering $\rho$ yields fine-grained, smooth control over the fractional bit-width, accommodating diverse deployment budgets.
Then, through super-group allocation and matrix multiplication $\smash{\mathbf{Y}=\mathbf{W}\mathbf{X}}$, each output element $\smash{Y_{i,l}}$ is computed as
\begin{equation} \label{eq:mixed-precision-forward}
Y_{i,l}
= \mathbf{W}^{G}_{i,\cdot}\,\mathbf{X}^{G}_{\cdot,l}
= \sum_{j=0}^{J-1}
  \underbrace{s^W_{i,j}\cdot s^X_{j,l}}_{{\textcircled{\scriptsize 1}}}
  \;\cdot\;
  \underbrace{Q_{n\text{-bit}}(\mathbf{W}^{G}_{i,j})^\top
              Q_{8\text{-bit}}(\mathbf{X}^{G}_{j,l})}_%
             {{\textcircled{\scriptsize 2}}}\ ,
  \quad n \in \{1.58, 4\} \ ,
\end{equation}
where $\smash{J=d_{\text{in}}/G}$ denotes the total number of groups along the input dimension $d_{\text{in}}$ for a given group size $G$. The group-wise vectors $\smash{\mathbf{W}^{G}_{i,j}}$ and $\smash{\mathbf{X}^{G}_{j,l}}$ represent the $j$-th group of the $i$-th weight output channel and the $l$-th activation token, respectively. The terms $s^W_{i,j}$ and $s^X_{j,l}$ are the corresponding 16-bit scaling factors, which are multiplied together to form the combined scaling factor \textcircled{\scriptsize 1}, and \textcircled{\scriptsize 2} is the low-bit integer dot product between the quantized weight and activation groups, where $n \in \{1.58, 4\}$ is strictly governed by the aforementioned super-group allocation. This factorization facilitates inference acceleration, as the integer arithmetic in~\textcircled{\scriptsize 2} can be offloaded to efficient kernels.

Unlike PTQ methods that statically preserve sensitive output channels~\citep{lin2024awq,huang2025slimllm}, QAT and QAD continuously update parameters, causing the salient weights to shift correspondingly.
We provide a theoretical justification for super-group allocation in Appendix~\ref{apx:optimal-allocation-proof}.

\subsection{Layer-adaptive feature distillation}
\label{subsec:LAFD}

We propose LAFD to adaptively identify the most informative layers for each input using a structural-similarity-based importance metric computed from the teacher. 
LAFD is motivated by the observation that consecutive transformer layers do not contribute equally to the overall feature transformation~\citep{tenney2019bert}.
Our empirical analysis in Appendix~\ref{apx:pattern-of-features} reveals that these transformation patterns are highly domain-dependent, which indicates the limitations of heuristic fixed-layer selection.
We explicitly quantify this structural similarity using cosine similarity to assess representational transformation, as angular differences between contextual features can reflect semantic and structural changes~\citep{ethayarajh2019how}. 
To quantify this transformation across layers, we compute the mean cosine similarity between the outputs of adjacent teacher layers across the set of valid token positions $\mathcal{T}$ excluding padding,
\begin{equation} \label{eq:cosine_metric}
c_{l}
= \frac{1}{|\mathcal{T}|}\sum_{t \in \mathcal{T}}
  \cos\!\left(\mathbf{F}_{T,t}^{(l)},\;
              \mathbf{F}_{T,t}^{(l-1)}\right) \ ,
  \quad l = 1, 2, \ldots, L \ ,
\end{equation}
where \smash{$\mathbf{F}_{T,t}^{(l)} \in \mathbb{R}^d$} denotes the $d$-dimensional teacher features within a training batch at layer $l$ and position $t$, and \smash{$\mathbf{F}_{T,t}^{(0)}$} corresponds to the output of the embedding layer.
A low value of \smash{$c_l$} indicates that layer $l$ substantially transforms the representation direction. We select the $k$ layers with the lowest scores as the targets of feature distillation,
\begin{equation} \label{eq:layer_select}
\mathcal{S}
= \underset{\substack{S \subseteq \{1,\ldots,L\},\;\lvert S\rvert = k}}
           {\arg\min} \;
  \sum_{l \in S} c_{l} \ ,
\end{equation}
% 公式的意思是：获得 top-k 最小的 c_l 对应的层索引集合 A，作为最终的 layer set S
where $\mathcal{S}$ is the set containing selected layers. The feature distillation loss is then defined over this adaptively selected set $\mathcal{S}$ as
\begin{equation} \label{eq:method3}
\mathcal{L}_{\mathrm{feature}}
= \text{MSE}_{\mathrm{adaptive}}\!\left(\mathbf{F}_T\,\middle\|\,\mathbf{F}_S\right)
= \frac{1}{\lvert\mathcal{S}\rvert}
  \sum_{l \in \mathcal{S}}
  \frac{1}{\lvert\mathcal{T}\rvert \cdot d}
  \sum_{t \in \mathcal{T}}
  \left\|\,\mathbf{F}_{T,t}^{(l)} - \mathbf{F}_{S,t}^{(l)}\right\|_{2}^{2} \ ,
\end{equation}
where \smash{$\mathbf{F}_{S,t}^{(l)}$} is the corresponding student features.
By restricting the feature distillation loss to $\mathcal{S}$, LAFD concentrates the gradient signal on layers with the largest representational changes. This helps reduce the propagation and amplification of quantization errors through subsequent nonlinear computations.
Through leveraging structural similarity scores, LAFD provides input-adaptive layer supervision while avoiding the prohibitive cost of searching over layer combinations.

% ---------------------------------------------------------

\subsection{Entropy-aware KL divergence}
\label{subsec:EAKLD}

In logit distillation, the KLD is used to align the student distribution $\smash{P_S}$ with the teacher distribution $\smash{P_T}$.
The forward KLD $\smash{\mathcal{D}_{\mathrm{KL}}(P_T \| P_S)}$ is zero-avoiding, inducing mode-covering behavior, while the reverse KLD $\smash{\mathcal{D}_{\mathrm{KL}}(P_S \| P_T)}$ is zero-forcing, inducing mode-seeking behavior~\citep{wu2025rethinking}.
CAKLD, introduced in BitDistiller~\citep{du2024bitdistiller}, balances these divergences using teacher confidence, but this metric leads to severe mismatch errors in both human-annotated and synthetic corpora, as demonstrated in Appendix~\ref{apx:teacher-confidence}, thereby restricting data diversity. To overcome this, we propose EAKLD, which interpolates between the two objectives using a batch-level mixing coefficient $\lambda$ derived from the teacher's output entropy. The logit distillation loss and the mixing coefficient are defined as
\begin{equation} \label{eq:eakld}
\begin{split}
\mathcal{L}_{\mathrm{logit}} = \mathcal{D}_{\mathrm{EAKLD}}\!\left(
    P_T
    \,\middle\|\,
    P_S
\right) &= 
\lambda \, \underbrace{ \mathcal{D}_{\mathrm{KL}}\!\left(
    P_T
    \,\middle\|\,
    P_S
\right)}_{\mathrm{forward\ KLD}} 
+ 
(1-\lambda) \, \underbrace{ \mathcal{D}_{\mathrm{KL}}\!\left(
    P_S
    \,\middle\|\,
    P_T
\right)}_{\mathrm{reverse\ KLD}} \ , \\
\text{with}\quad {\lambda} &= \mathbb{E}_{(x,y)\sim\mathbb{D}}\!\left[
  \frac{1}{|y|}
  \sum_{i=1}^{|y|}
  \frac{
    \min\!\Big(
      H\big(P_T(y_i|x, y_{<i})\big),\;
      \log K
    \Big)
  }{\log K}
\right] \ ,
\end{split}
\end{equation}
where $\mathbb{D}$ is the data within a training batch, $|y|$ is the number of tokens in the response sequence, and $H(P_T(x, y_{<i}))$ denotes the entropy of the teacher's predictive distribution at position $i$ conditioned on the input $x$ and the preceding tokens $y_{<i}$. The entropy is formulated as
\begin{equation}
    H\big(P_T(y_i|x, y_{<i})\big) = -\sum_{v \in \mathcal{V}} P_T(v \mid x, y_{<i}) \log P_T(v \mid x, y_{<i}) \ ,
\end{equation}
where $\mathcal{V}$ is the vocabulary set, and the denominator $\log K$ represents the maximum entropy of a $K$-uniform distribution.
When the teacher disperses probability evenly among candidates, the entropy increases, causing $\lambda$ to rise and adaptively strengthening the forward KLD, thereby encouraging mode-covering.
Conversely, when the teacher places high confidence among candidates, yielding a small $H(P_T)$, $\lambda$ decays, thereby prioritizing the reverse KLD for precise mode-seeking. 
Furthermore, tuning the hyperparameter $K$ deterministically alters the upper-bound entropy, serving as the primary control mechanism to adjust the entire dataset's aggregate tendency toward either divergence strategy.

By adapting logit distillation exclusively from entropy, EAKLD captures the full shape of the teacher's uncertainty and avoids the need for internally distilled labels, supporting training corpora that comprise both human-annotated and externally distilled datasets.

% ---------------------------------------------------------

\section{Experiments}
\label{sec:experiments}

In this section, we conduct comprehensive experiments to validate the effectiveness and efficiency of the proposed \textsc{EdgeRazor} framework along with its three modules.

\subsection{Experimental setup}
\label{sec:experimental-configurations}
We apply the \textsc{EdgeRazor} framework to four models: MobileLLM-ParetoQ-350M-BF16 (Mo-bileLLM-350M)~\citep{liu2025paretoq} representing the base LLMs, Qwen3-0.6B and Qwen3-1.7B~\citep{yang2025qwen3} \zhangsqzhangsh{indicating} the instruction-tuned LLMs, and Qwen2.5-Omni-7B~\citep{xu2025qwen2} \zhangsqzhangsh{referring to} the multimodal LLMs.
% 这里的 (Mo-bileLLM-350M) 需要注意格式

\paragraph{Datasets.}
We train the three text-only LLMs on a mixture of human-annotated and externally distilled instruction data~\citep{li2025infinity, zhao2025amdistill}, combined with the training splits of commonsense tasks~\citep{bisk2020piqa, clark2019boolq, clark2018arc, lin2022truthfulqa, mihaylov2018openbookqa, sakaguchi2021winogrande, sap2019socialiqa, zellers2019hellaswag}. All synthetic samples are generated by external high-quality LLMs rather than their corresponding teachers.
In addition, the multimodal LLM is trained on 10K TGIF~\citep{li2016tgif} clips re-encoded at 30\,FPS, with video-understanding responses distilled from the teacher.
Appendix~\ref{apx:training-data} details the specific data mixtures and usage for each model.

\begin{table}[!t]
\caption{Hyperparameters for \textsc{EdgeRazor}.}
\label{tab:hyperparameters}
\centering
\begin{tabular}{lcccccccc}
\toprule
\textbf{Models} & $\boldsymbol{\rho}$ & \textbf{Bit-widths} & \textbf{LRs} & \textbf{Training} & $\boldsymbol{\alpha_\mathrm{task}}$ & $\boldsymbol{\alpha_\mathrm{feature}}$ & $\boldsymbol{\alpha_\mathrm{logit}}$ \\
\midrule
\multirow{4}{*}{\textbf{MobileLLM-350M}}
 & $1$   & 4.00 & \multirow{4}{*}{$2\!\times\!10^{-5}$} & 2 epochs & 0.50 & 0.10 & 2.0 \\
 & $1/2$ & 2.79 & & 4 epochs & 0.50 & 0.10 & 2.0 \\
 & $1/8$ & 1.88 & & 5 epochs & 0.20 & 1.00 & 4.0 \\
 & $0$   & 1.58 & & 5 epochs & 0.20 & 1.00 & 4.0 \\
\midrule
\multirow{4}{*}{\textbf{Qwen3-0.6B}}
 & $1$   & 4.00 & \multirow{4}{*}{$2\!\times\!10^{-5}$} & 2k steps & 0.05 & 0.50 & 2.0 \\
 & $1/2$ & 2.79 & & 1 epoch  & 0.10 & 0.10 & 2.0 \\
 & $1/8$ & 1.88 & & 1 epoch  & 0.10 & 0.10 & 2.0 \\
 & $0$   & 1.58 & & 1 epoch  & 0.10 & 0.10 & 2.0 \\
\midrule
\multirow{4}{*}{\textbf{Qwen3-1.7B}}
 & $1$   & 4.00 & \multirow{4}{*}{$2\!\times\!10^{-5}$} & 2k steps & 0.05 & 0.50 & 2.0 \\
 & $1/2$ & 2.79 & & 2 epochs & 0.10 & 0.10 & 2.0 \\
 & $1/8$ & 1.88 & & 2 epochs & 0.10 & 0.10 & 2.0 \\
 & $0$   & 1.58 & & 2 epochs & 0.10 & 0.10 & 2.0 \\
\midrule
\textbf{Qwen2.5-Omni-7B} & $1$   & 4.00 & $5\!\times\!10^{-6}$ & 2 epochs & 0.10 & 0.20 & 2.0 \\
\bottomrule
\end{tabular}
\end{table}

\paragraph{Hyperparameters.}
All low-bit LLMs are trained on 8$\times$NVIDIA~A100-80\,GB GPUs. We apply per-group symmetric quantization, setting the group sizes to 256, 64, and 32 for Qwen3, MobileLLM, and Qwen2.5. The embedding and \texttt{lm\_head} layers are kept at 4-bit. For all configurations, we use the AdamW optimizer, and set $\beta{=}2.0$, $\epsilon{=}10^{-5}$, $k{=}3$ in LAFD, and $K{=}16$ in EAKLD. For training schedules, Qwen3-0.6B and 1.7B employ a constant learning rate with a 0.05 warmup ratio and batch sizes of 1024 and 1536. MobileLLM-350M and Qwen2.5-Omni-7B utilize a cosine schedule with a 0.01 warmup ratio and batch sizes of 1920 and 64. Additional details are provided in Table~\ref{tab:hyperparameters}.

\paragraph{Baselines.} We compare \textsc{EdgeRazor} against leading PTQ and QAT baselines, including GPTQ~\citep{frantar2023gptq}, AWQ~\citep{lin2024awq}, AQLM~\citep{egiazarian2024aqlm}, QTIP~\citep{tseng2024qtip}, Slim-LLM+~\citep{huang2025slimllm}, AutoRound~\citep{cheng2024autoround}, Q-Palette~\citep{lee2025qpalette}, EfficientQAT~\citep{chen2024efficientqat}, ParetoQ~\citep{liu2025paretoq}, LQER~\citep{zhang2024lqer}, OmniQuant~\citep{shao2024omniquant}, ABQ-LLM~\citep{li2025abq}, SpinQuant~\citep{liu2025spinquant}, and FlatQuant~\citep{sun2025flatquant}. The baseline group sizes are 64 for MobileLLM and 128 for the other LLMs. We also provide evaluations with per-task results against more baselines in Appendix~\ref{apx:effectiveness}.
For the ablation studies, we compare three aspects: the super-group allocation in SQMP depicted in Figure~\ref{fig:super-group-allocation} against the stacked allocation in Figure~\ref{fig:stacked-allocation};
LAFD against conventional feature distillation;
and EAKLD in logit distillation against forward KLD and CAKLD introduced in the QAD baseline BitDistiller~\citep{du2024bitdistiller}.

\paragraph{Evaluation.} Table~\ref{tab:benchmarks} summarizes the evaluation protocols. We prioritize domain-specific tasks over generic perplexity to reflect the concrete reasoning and generation capabilities of LLMs. The text LLMs are benchmarked across diverse tasks with random seed 42, including commonsense reasoning~\citep{bisk2020piqa,clark2019boolq,clark2018arc,sakaguchi2021winogrande,sap2019socialiqa,zellers2019hellaswag}, reading comprehension~\citep{mihaylov2018openbookqa}, trustworthiness~\citep{hendrycks2021ethics}, truthfulness~\citep{lin2022truthfulqa}, knowledge~\citep{hendrycks2021mmlu}, instruction following~\citep{zhou2023ifeval}, mathematics~\citep{cobbe2021gsm8k}, and coding~\citep{chen2021humaneval}, using the \href{https://github.com/EleutherAI/lm-evaluation-harness/tree/v0.4.9.1}{\texttt{lm\_eval} v0.4.9.1}. Qwen2.5-Omni-7B is evaluated on two video understanding tasks~\citep{fu2024videomme,zhou2024mlvu} using the \href{https://github.com/evolvinglmms-lab/lmms-eval/tree/v0.5}{\texttt{lmms\_eval} v0.5}. To evaluate deployment efficiency, we benchmark on an Apple M4 Pro chip using \texttt{llama-bench} in \href{https://github.com/ggml-org/llama.cpp/tree/b6300}{\texttt{llama.cpp} b6300} under 100 repetitions, with additional details in Appendix~\ref{apx:efficiency}.

\subsection{Main results}

\begin{figure*}[!ht]
    \centering
    \includegraphics[width=1\textwidth]{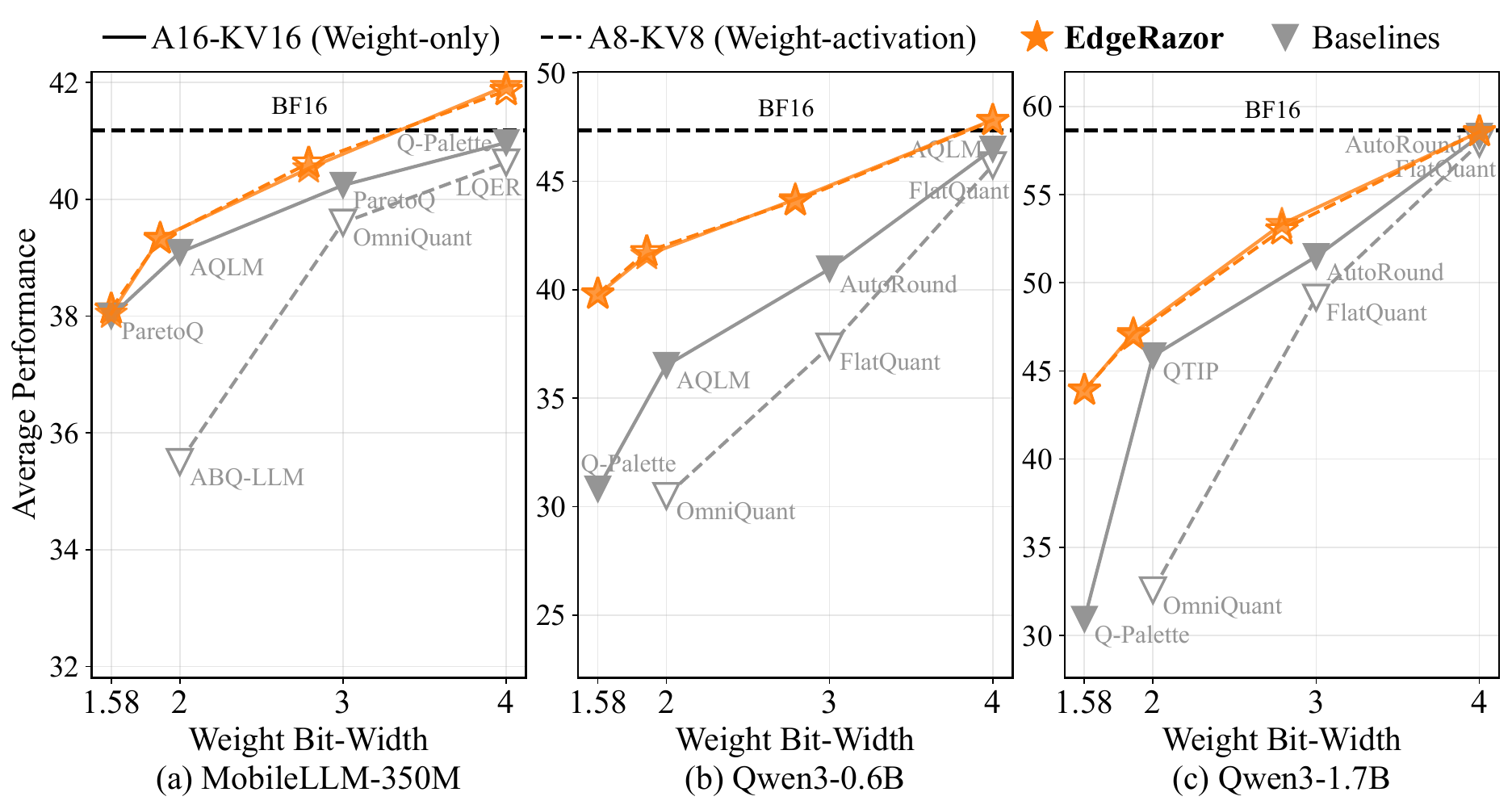}
    \caption{Performance comparison of \textsc{EdgeRazor} and state-of-the-art baselines at each bit-width.}
    \label{fig:edgerazor-baselines}
\end{figure*}

\begin{figure*}[!ht]
    \centering
    \includegraphics[width=1\textwidth]{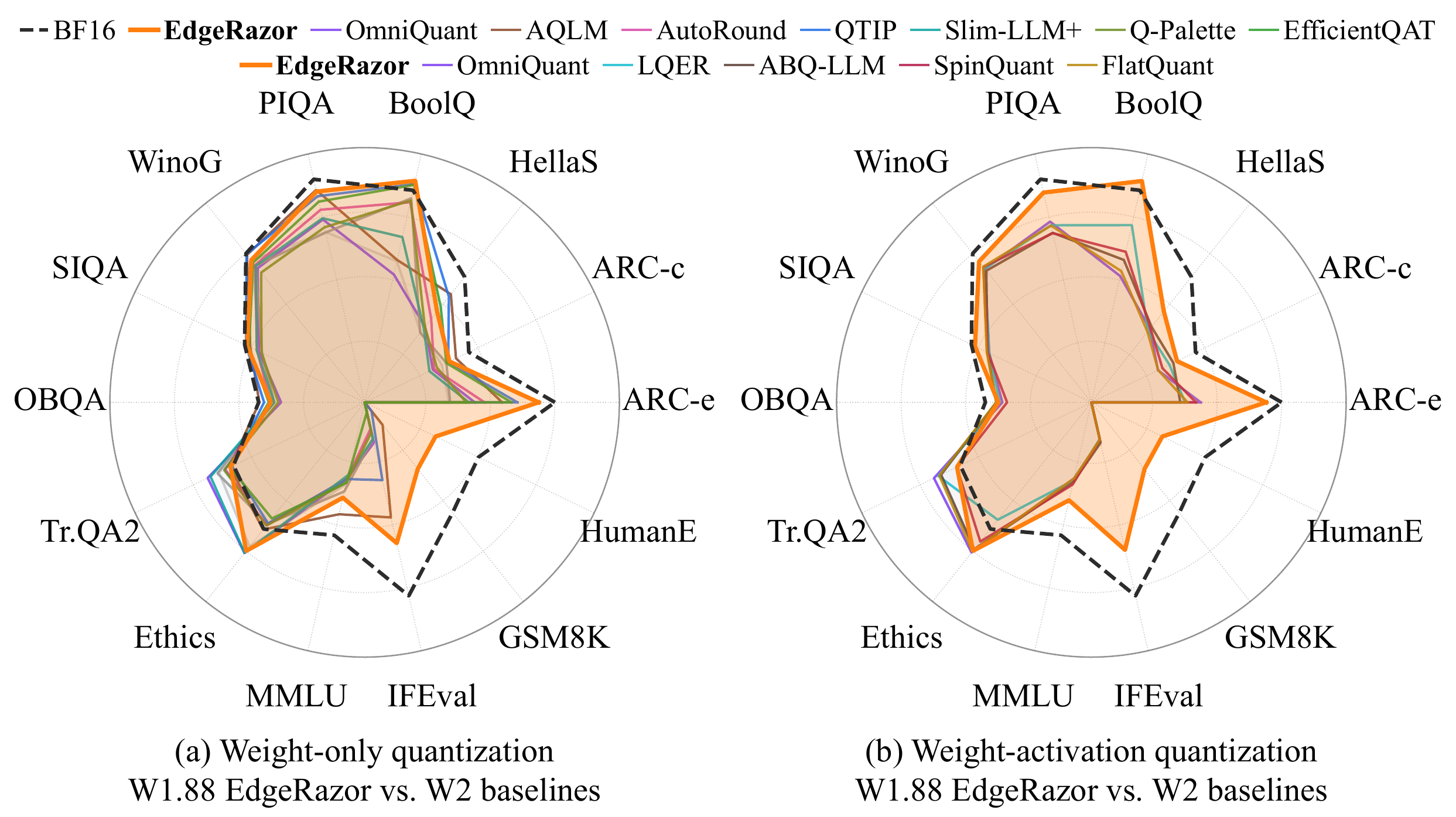}
    \caption{Performance comparison of 1.88-bit \textsc{EdgeRazor} and 2-bit baselines on Qwen3-0.6B.}
    \label{fig:qwen3-0.6b-2bit-perf-comparison}
\end{figure*}

\begin{figure*}[!t]
    \centering
    \vspace{-10pt}
    \includegraphics[width=1\textwidth]{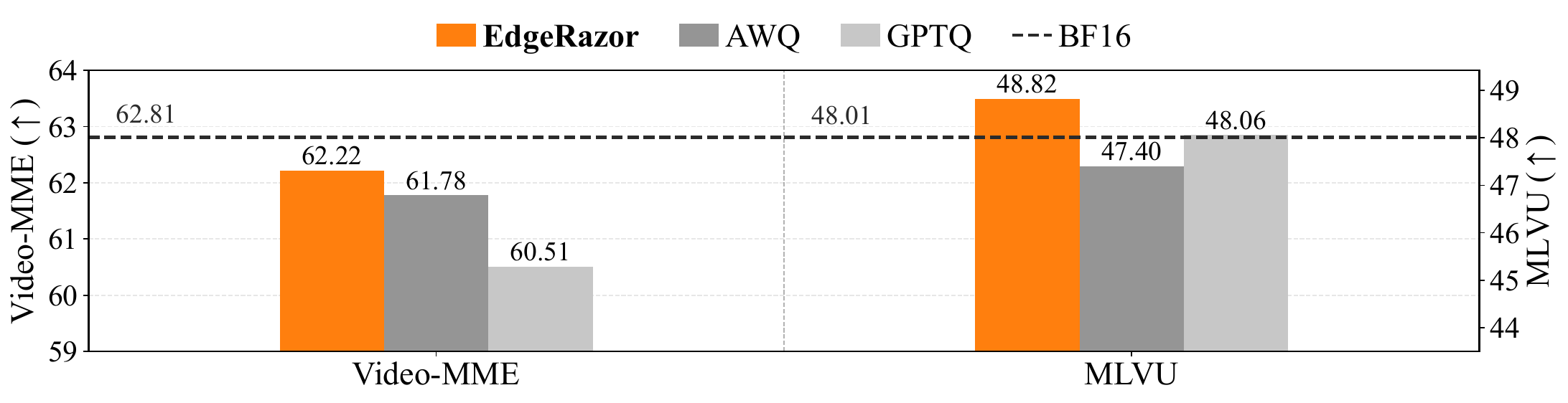}
    \caption{Performance comparison of 4-bit \textsc{EdgeRazor} and strong baselines on Qwen2.5-Omni-7B.}
    \label{fig:qwen2.5-perf-comparison}
    \vspace{-10pt}
\end{figure*}

\begin{figure*}[!t]
    \centering
    \includegraphics[width=1\textwidth]{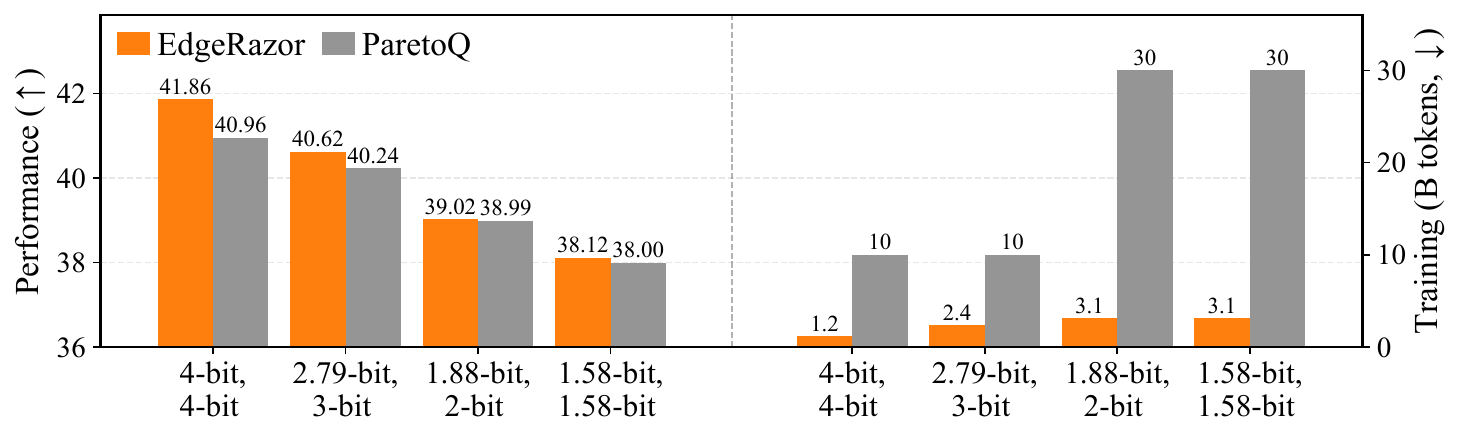}
    \caption{Average performance and training budgets of \textsc{EdgeRazor} and ParetoQ on MobileLLM-350M. The training budgets are reported in tokens consumed during training.}
    \label{fig:budget-comparison}
\end{figure*}

We conduct comprehensive experiments on downstream tasks listed in Appendix~\ref{apx:eval-protocols}.
Figure~\ref{fig:edgerazor-baselines} reports the average performance of \textsc{EdgeRazor}, PTQ, and QAT state-of-the-art baselines on MobileLLM-350M, Qwen3-0.6B, and Qwen3-1.7B under both weight-only and weight-activation quantization.
Table~\ref{tab:qwen3-0.6b} provides detailed performance on Qwen3-0.6B at every bit-width.
Figure~\ref{fig:qwen3-0.6b-2bit-perf-comparison} visualizes per-task performance at the challenging 1.88-bit setting.
Figure~\ref{fig:qwen2.5-perf-comparison} extends the evaluation to the multimodal LLM Qwen2.5-Omni-7B.
Figure~\ref{fig:budget-comparison} compares the average performance and training budgets of \textsc{EdgeRazor} and the leading QAT baseline ParetoQ. Per-task results are given in Appendix~\ref{apx:effectiveness}.

We make eight key observations based on the results:
(1) In Figure~\ref{fig:edgerazor-baselines}, \textsc{EdgeRazor} consistently surpasses the strongest baselines across all bit-widths on the three text-only LLMs, and the margin widens as the bit-width decreases;
(2) In Figure~\ref{fig:edgerazor-baselines}, the weight-activation baseline curve lies noticeably below the weight-only one, whereas the two \textsc{EdgeRazor} curves nearly coincide, and the gap between \textsc{EdgeRazor} and baselines is larger under weight-activation quantization;
(3) In Table~\ref{tab:qwen3-0.6b}, on Qwen3-0.6B, at 4-bit, \textsc{EdgeRazor} leads AQLM by 1.35 and FlatQuant by 2.06 points; at 2.79-bit, it achieves gains of 3.21 and 6.72 points over 3-bit AutoRound and FlatQuant; at 1.88-bit, it improves upon 2-bit AQLM and OmniQuant by 5.09 and 11.27 points; at 1.58-bit, it obtains an 8.96-point gain over the 1.75-bit Q-Palette;
(4) In Table~\ref{tab:qwen3-0.6b}, the 1.88-bit \textsc{EdgeRazor} outperforms the strongest 3-bit baselines AutoRound and FlatQuant by 0.64 and 4.38 points, and the 1.58-bit \textsc{EdgeRazor} exceeds the strongest 2-bit weight-only baseline AQLM by 3.26 points and the strongest 3-bit weight-activation baseline FlatQuant by 2.43 points;
(5) In Table~\ref{tab:qwen3-0.6b}, \textsc{EdgeRazor} presents superiority over mixed-precision methods like Q-Palette and Slim-LLM+;
(6) In Figure~\ref{fig:qwen3-0.6b-2bit-perf-comparison}, most 2-bit baselines collapse to near-zero scores on the reasoning-intensive GSM8K and code-generation HumanEval tasks, while \textsc{EdgeRazor} retains a clear advantage on both;
(7) In Figure~\ref{fig:qwen2.5-perf-comparison}, on the multimodal LLM, Qwen2.5-Omni-7B, at 4-bit weight-only quantization, \textsc{EdgeRazor} with an additional 4-bit vision encoder surpasses AWQ by 0.44 on Video-MME and 1.42 on MLVU;
(8) In Figure~\ref{fig:budget-comparison}, on MobileLLM-350M, \textsc{EdgeRazor} trained via MPQAD outperforms ParetoQ trained by QAT finetuning at every bit-width, while requiring a $4$--$10\times$ lower training budget.
% highlighting its superior efficiency--accuracy trade-off.

Observations (1)--(4) demonstrate that \textsc{EdgeRazor} consistently achieves superior performance, with particularly pronounced gains in the ultra-low-bit regime.
Observation (5) confirms the effectiveness of our mixed-precision module, SQMP.
Observation (6) and Appendix~\ref{apx:capability-low-bit-llms} indicate that \textsc{EdgeRazor} preserves complex reasoning and generation capabilities under ultra-low-bit settings.
Observation (7) shows that the advantages of \textsc{EdgeRazor} extend beyond text-only LLMs.
Observation (8) highlights the superior efficiency-performance trade-off of \textsc{EdgeRazor}.
Appendix~\ref{apx:generalization-held-out} further validates its robust generalization on held-out benchmarks.
These results establish the effectiveness of \textsc{EdgeRazor}.

\begin{table}[!t]
    \centering
    \caption{Performance comparison of \textsc{EdgeRazor} and strong baselines on Qwen3-0.6B.}
    \label{tab:qwen3-0.6b}
    \setlength{\tabcolsep}{2.5pt}
    % \small
    %====================== (a) Weight-only ======================
    \begin{subtable}[t]{0.49\linewidth}
    \centering
    \caption{Weight-only quantization.}
    \label{tab:qwen3-0.6b-w}
    \begin{tabular}{lccr}
      \toprule
      \textbf{Method} & \textbf{W-A-KV} & \textbf{Avg. ($\uparrow$)} & \textbf{$\Delta_{\text{BF16}}$ ($\uparrow$)} \\
      \midrule
      % ----- 4-bit group -----
      AQLM         & 4-16-16 & 46.48 & -0.87 \\
      AutoRound    & 4-16-16 & 45.75 & -1.60 \\
      Q-Palette    & 4-16-16 & 40.97 & -6.38 \\
      EfficientQAT & 4-16-16 & 44.07 & -3.28 \\
      \rowcolor{gray!20}
      \textbf{\textsc{EdgeRazor}} & \textbf{4-16-16} & \textbf{47.83} & \textbf{+0.48} \\
      \midrule
      % ----- 3-bit group -----
      AutoRound    & 3-16-16    & 40.96 & -6.39  \\
      Slim-LLM+    & 3-16-16    & 33.95 & -13.40 \\
      Q-Palette    & 3.25-16-16 & 37.55 & -9.80  \\
      EfficientQAT & 3-16-16    & 39.92 & -7.43  \\
      \rowcolor{gray!20}
      \textbf{\textsc{EdgeRazor}} & \textbf{2.79-16-16} & \textbf{44.17} & \textbf{-3.18} \\
      \midrule
      % ----- 2-bit group -----
      AQLM         & 2-16-16 & 36.51 & -10.84 \\
      QTIP         & 2-16-16 & 35.94 & -11.41 \\
      Slim-LLM+    & 2-16-16 & 30.54 & -16.81 \\
      Q-Palette    & 2-16-16 & 30.66 & -16.69 \\
      EfficientQAT & 2-16-16 & 33.27 & -14.08 \\
      \rowcolor{gray!20}
      \textbf{\textsc{EdgeRazor}} & \textbf{1.88-16-16} & \textbf{41.60} & \textbf{-5.75} \\
      \midrule
      % ----- <2-bit group -----
      Q-Palette    & 1.75-16-16 & 30.81 & -16.54 \\
      \rowcolor{gray!20}
      \textbf{\textsc{EdgeRazor}} & \textbf{1.58-16-16} & \textbf{39.77} & \textbf{-7.58} \\
      \bottomrule
    \end{tabular}
    \end{subtable}
    \hfill
    %==================== (b) Weight-activation ====================
    \begin{subtable}[t]{0.49\linewidth}
      \centering
      \caption{Weight-activation quantization.}
      \label{tab:qwen3-0.6b-wa}
      \begin{tabular}{lccr}
        \toprule
        \textbf{Method} & \textbf{W-A-KV} & \textbf{Avg. ($\uparrow$)} & \textbf{$\Delta_{\text{BF16}}$ ($\uparrow$)} \\
        \midrule
        % ----- 4-bit group -----
        OmniQuant & 4-8-8 & 37.27 & -10.08 \\
        LQER      & 4-8-8 & 45.31 & -2.04  \\
        SpinQuant & 4-8-8 & 41.27 & -6.08  \\
        FlatQuant & 4-8-8 & 45.74 & -1.61  \\
        \rowcolor{gray!20}
        \textbf{\textsc{EdgeRazor}} & \textbf{4-8-8} & \textbf{47.80} & \textbf{+0.45} \\
        \midrule
        % ----- 3-bit group -----
        OmniQuant & 3-8-8 & 34.58 & -12.77 \\
        LQER      & 3-8-8 & 36.46 & -10.89 \\
        SpinQuant & 3-8-8 & 34.93 & -12.42 \\
        FlatQuant & 3-8-8 & 37.38 & -9.97  \\
        \rowcolor{gray!20}
        \textbf{\textsc{EdgeRazor}} & \textbf{2.79-8-8} & \textbf{44.10} & \textbf{-3.25} \\
        \midrule
        % ----- 2-bit group -----
        OmniQuant & 2-8-8 & 30.49 & -16.86 \\
        LQER      & 2-8-8 & 30.46 & -16.89 \\
        SpinQuant & 2-8-8 & 30.04 & -17.31 \\
        FlatQuant & 2-8-8 & 30.23 & -17.12 \\
        \rowcolor{gray!20}
        \textbf{\textsc{EdgeRazor}} & \textbf{1.88-8-8} & \textbf{41.76} & \textbf{-5.59} \\
        \midrule
        % ----- <2-bit group -----
        \rowcolor{gray!20}
        \textbf{\textsc{EdgeRazor}} & \textbf{1.58-8-8} & \textbf{39.81} & \textbf{-7.54} \\
        \bottomrule
      \end{tabular}
    \end{subtable}
\end{table}

% ----------------------------------------------------------

\subsection{Ablation studies}

\begin{table}[!ht]
  \centering
  \caption{Configurations for the ablation studies of \textsc{EdgeRazor}.}
  \label{tab:ablation_studies_config}
  \setlength{\tabcolsep}{1.5pt}
  \begin{tabular}{lcccccccc}
    \toprule
    \multirow{2}{*}{\textbf{Methods}} & \multicolumn{2}{c}{\textbf{Allocation}} & \multicolumn{2}{c}{\textbf{Feature distillation}} & \multicolumn{3}{c}{\textbf{Logit distillation}} \\
    \cmidrule(lr){2-3} \cmidrule(lr){4-5} \cmidrule(lr){6-8}
    & \textbf{Super-group} & \textbf{Stacked} & \textbf{Adaptive} & \textbf{Fixed} & \textbf{EAKLD} & \textbf{CAKLD} & \textbf{FKLD} \\
    \midrule
    \rowcolor{gray!20}
    \textbf{\textsc{EdgeRazor}$_\text{SG+A+E}$} & \checkmark & \texttimes & \checkmark & \texttimes     & \checkmark & \texttimes     & \texttimes     \\
    \textbf{\textsc{EdgeRazor}$_\text{ST+A+E}$} & \texttimes & \checkmark & \checkmark & \texttimes     & \checkmark & \texttimes     & \texttimes     \\
    \textbf{\textsc{EdgeRazor}$_\text{SG+A+C}$} & \checkmark & \texttimes & \checkmark & \texttimes     & \texttimes     & \checkmark & \texttimes     \\
    \textbf{\textsc{EdgeRazor}$_\text{SG+F+E}$} & \checkmark & \texttimes & \texttimes     & \checkmark & \checkmark & \texttimes     & \texttimes     \\
    \textbf{\textsc{EdgeRazor}$_\text{SG+F+F}$} & \checkmark & \texttimes & \texttimes     & \checkmark & \texttimes     & \texttimes     & \checkmark \\
    \bottomrule
  \end{tabular}
\end{table}

\begin{table}[!ht]
  \centering
  \caption{Ablation studies of \textsc{EdgeRazor} on Qwen3-0.6B.}
  \label{tab:ablation_studies}
  \begin{tabular}{lcccccccc}
    \toprule
    \textbf{Methods} & \textbf{W} & \textbf{A} & \textbf{KV} & \textbf{Average} & \textbf{W} & \textbf{A} & \textbf{KV} & \textbf{Average} \\
    \midrule
    \rowcolor{gray!20}
    \textbf{\textsc{EdgeRazor}$_\text{SG+A+E}$} & 2.79 & 16 & 16 & \textbf{44.17} & 2.79 & 8 & 8 & \textbf{44.10} \\
    \textbf{\textsc{EdgeRazor}$_\text{ST+A+E}$} & 2.79 & 16 & 16 & 43.26 & 2.79 & 8 & 8 & 43.08 \\
    \midrule
    \rowcolor{gray!20}
    \textbf{\textsc{EdgeRazor}$_\text{SG+A+E}$} & 2.19 & 16 & 16 & \textbf{40.71} & 2.19 & 8 & 8 & \textbf{40.14} \\
    \textbf{\textsc{EdgeRazor}$_\text{SG+A+C}$} & 2.19 & 16 & 16 & 39.59 & 2.19 & 8 & 8 & 39.61 \\
    \textbf{\textsc{EdgeRazor}$_\text{SG+F+E}$} & 2.19 & 16 & 16 & 40.27 & 2.19 & 8 & 8 & 39.93 \\
    \textbf{\textsc{EdgeRazor}$_\text{SG+F+F}$} & 2.19 & 16 & 16 & 39.25 & 2.19 & 8 & 8 & 39.51 \\
    \midrule
    \rowcolor{gray!20}
    \textbf{\textsc{EdgeRazor}$_\text{SG+A+E}$} & 1.88 & 16 & 16 & \textbf{41.60} & 1.88 & 8 & 8 & \textbf{41.76} \\
    \textbf{\textsc{EdgeRazor}$_\text{SG+A+C}$} & 1.88 & 16 & 16 & 40.95 & 1.88 & 8 & 8 & 40.85 \\
    \textbf{\textsc{EdgeRazor}$_\text{SG+F+E}$} & 1.88 & 16 & 16 & 40.40 & 1.88 & 8 & 8 & 40.24 \\
    \textbf{\textsc{EdgeRazor}$_\text{SG+F+F}$} & 1.88 & 16 & 16 & 39.83 & 1.88 & 8 & 8 & 39.70 \\
    \bottomrule
  \end{tabular}
\end{table}

To isolate the sources of these performance gains, we conduct detailed ablation studies on the three proposed modules: SQMP, LAFD, and EAKLD. We employ five configurations in Table~\ref{tab:ablation_studies_config} and evaluate their corresponding performance in Table~\ref{tab:ablation_studies}. Detailed settings and per-task results are provided in Appendix~\ref{apx:detailed-ablation}.
There are four key findings:
(1) At 2.79-bit, \textsc{EdgeRazor}$_\text{SG+A+E}$ consistently outperforms \textsc{EdgeRazor}$_\text{ST+A+E}$, yielding an improvement of 0.91 and 1.02 points under weight-only and weight-activation quantization setups, respectively;
(2) Replacing EAKLD with CAKLD leads to consistent performance drops, where \textsc{EdgeRazor}$_\text{SG+A+E}$ improves over \textsc{EdgeRazor}$_\text{SG+A+C}$ by 1.12 and 0.53 points at 2.19-bit, and by 0.65 and 0.91 points at 1.88-bit;
(3) Reverting from EAKLD to standard forward KLD similarly degrades performance, with \textsc{EdgeRazor}$_\text{SG+F+E}$ exceeding \textsc{EdgeRazor}$_\text{SG+F+F}$ by 1.02 and 0.42 points at 2.19-bit, and by 0.57 and 0.54 points at 1.88-bit;
(4) Enabling LAFD over heuristic fixed-layer selection yields consistent improvements, as evidenced by \textsc{EdgeRazor}$_\text{SG+A+E}$ improving upon \textsc{EdgeRazor}$_\text{SG+F+E}$ by 0.44 and 0.21 points at 2.19-bit, and by 1.20 and 1.52 points at 1.88-bit.

Finding (1) validates the advantage of super-group allocation over the stacked allocation utilized in the SQMP module. Findings (2) and (3) confirm the superiority of the EAKLD module over alternative distillation criteria, such as CAKLD and forward KLD. Finding (4) demonstrates the efficacy of the LAFD module relative to heuristic methods.
These results indicate that each module contributes to the overall performance.

% ----------------------------------------------------------

\subsection{Compression and acceleration}

\begin{figure*}[!ht]
    \centering
    \includegraphics[width=1\textwidth]{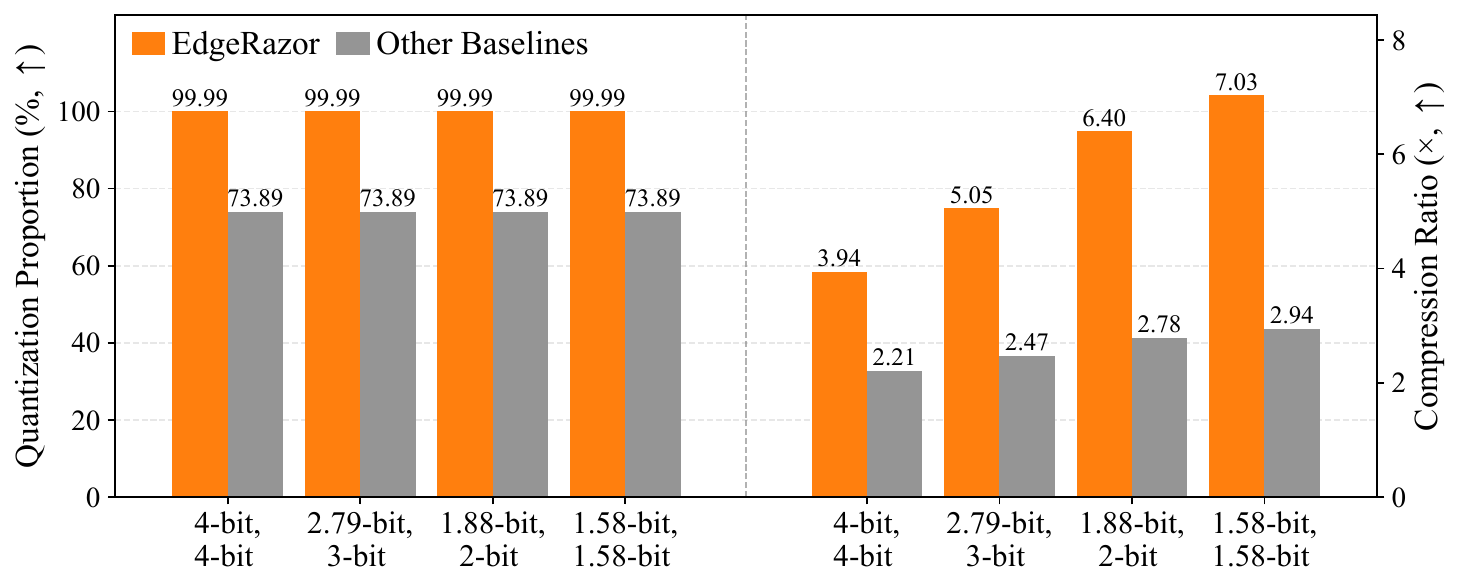}
    \caption{Efficiency comparison of \textsc{EdgeRazor} and other baselines at each bit-width.}
    \label{fig:compr}
\end{figure*}

\begin{figure*}[!ht]
    \centering
    \includegraphics[width=1\textwidth]{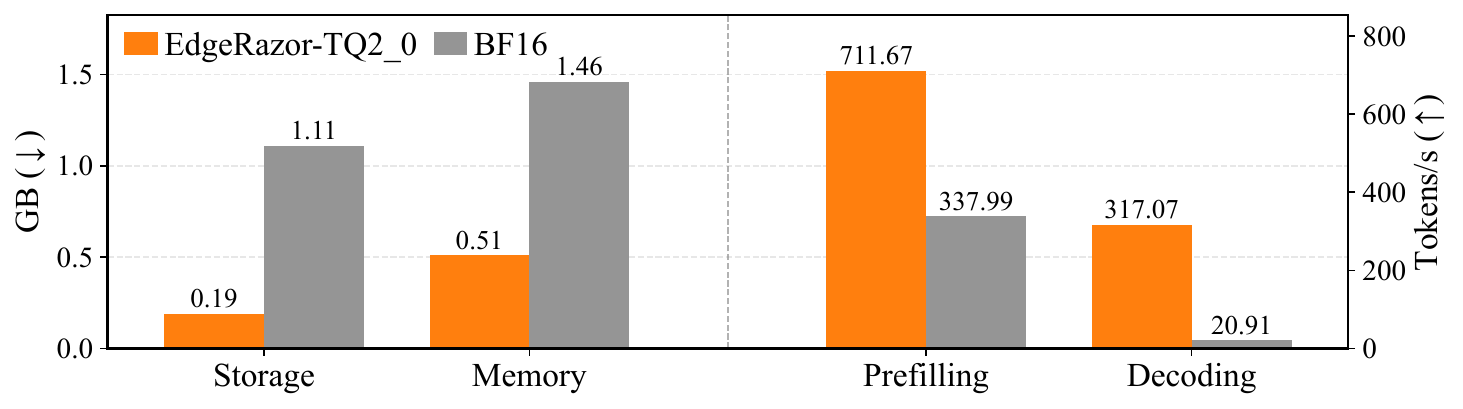}
    \caption{Efficiency comparison of deploying 1.58-bit and 4-bit Qwen3-0.6B via \texttt{llama.cpp}.}
    \label{fig:inference}
\end{figure*}

Beyond performance, deployment efficiency is important for edge applications.
We evaluate Qwen3-0.6B and outline two outcomes:
(1) By quantizing the embedding and \texttt{lm\_head} alongside decoder layers, \textsc{EdgeRazor} with 256 group\_size achieves a 99.99\% quantization proportion, maximizing compression ratios across all bit-widths over existing per-group quantization methods with 128 group\_size in Figure~\ref{fig:compr};
(2) At 1.58-bit \texttt{TQ2\_0} precision type, \textsc{EdgeRazor} reduces storage from 1.11\,GB to 0.19\,GB and memory from 1.46\,GB to 0.51\,GB, while accelerating prefilling and decoding speeds from 337.99 and 20.91 tokens/s to 711.67 and 317.03 tokens/s, achieving 2.11$\times$ and 15.16$\times$ acceleration over the BF16 baseline in Figure~\ref{fig:inference}.

Outcome (1) highlights the substantial compression achieved by our framework across all bit-widths. Outcome (2) confirms that this compression translates directly into reductions in resource and on-device acceleration. Appendix~\ref{apx:efficiency} further details these efficiency metrics for diverse LLMs, and the deployment benchmarks are conducted on both Apple M4 Pro and i9-14900K chips.
These results reveal that \textsc{EdgeRazor} provides an efficient solution for deploying low-bit LLMs.

%%%%%%%%%%%%%%%%%%%%%%%%%%%%%%%%%%%%%%%%%%%%%%%%%%%%%%%%%%%%
\section{Conclusions}
\label{sec:conclusion}

In this paper, we proposed \textsc{EdgeRazor}, a lightweight framework for compressing LLMs into flexible low-bit formats via MPQAD. It consisted of three novel modules: SQMP, LAFD, and EAKLD. Comprehensive evaluations across base, instruction-tuned, and multimodal LLMs demonstrated that \textsc{EdgeRazor} outperformed state-of-the-art baselines at all evaluated bit-widths in terms of domain-specific performance and training budget reduction. Practical benchmarks on \texttt{llama.cpp} demonstrated that \textsc{EdgeRazor} transformed aggressive quantization from a fragile trade-off into a resilient approach for deploying lightweight LLMs on resource-constrained hardware.

%%%%%%%%%%%%%%%%%%%%%%%%%%%%%%%%%%%%%%%%%%%%%%%%%%%%%%%%%%%%

\section{Acknowledgement}
Shao-Qun Zhang is the corresponding author, supported by the Natural Science Foundation of China (62406138) and the Natural Science Foundation of Jiangsu Province (BK20230782).
This research was supported by the Fundamental and Interdisciplinary Disciplines Breakthrough Plan of the Ministry of Education of China (No.\ JYB2025XDXM118). 

This research was performed during the academic cooperation between LAMDA and Microsoft AI, where the proposed \textsc{EdgeRazor} framework was evaluated and deployed within the internal systems.

%%%%%%%%%%%%%%%%%%%%%%%%%%%%%%%%%%%%%%%%%%%%%%%%%%%%%%%%%%%%

{
\small
\bibliographystyle{plain}
\bibliography{reference}
}

%%%%%%%%%%%%%%%%%%%%%%%%%%%%%%%%%%%%%%%%%%%%%%%%%%%%%%%%%%%%

\newpage
\appendix

%%%%%%%%%%%%%%%%%%%%%%%%%%%%%%%%%%%%%%%%%%%%%%%%%%%%%%%%%%%%

\section{Mixed-precision quantization}
\subsection{Quantization function for weights and activations}
\label{apx:quant-func}
In this section, we provide the per-group symmetric quantization for both weights and activations in our framework.
Let ${\mathbf{W}\in\mathbb{R}^{d_{\text{out}}\times d_{\text{in}}}}$ denote a weight matrix and ${\mathbf{X}\in\mathbb{R}^{d_{\text{in}}\times L}}$ denote the corresponding activation matrix, where $L$ is the sequence length.
Given a group size $G$, we partition $\mathbf{W}$ and $\mathbf{X}$ along the input dimension into ${J=d_{\text{in}}/G}$ groups per output channel to obtain ${\mathbf{W}^{G}\in\mathbb{R}^{d_{\text{out}}\times J \times G}}$ and ${\mathbf{X}^{G}\in\mathbb{R}^{J \times G \times L}}$.
The $j$-th group of the $i$-th output channel is defined as ${\mathbf{W}^{G}_{i,j}=\mathbf{W}[i,\;jG:(j{+}1)G]\in\mathbb{R}^{G}}$, with ${W^{G}_{i,j,k}}$ denoting its $k$-th element. Similarly, the $j$-th input-channel group for token~$l$ is defined as ${\mathbf{X}^{G}_{j,l}=\mathbf{X}[jG:(j{+}1)G,\;l] \in\mathbb{R}^{G}}$. Each group is independently quantized to $n$-bit through a quantization function applicable to both weights and activations,
\begin{equation} \label{eq:quant_func}
Q_{n\text{-bit}}(\mathbf{W}^{G}_{i,j}) = \begin{cases}
  \text{clip}\!\left(\left\lfloor\dfrac{\mathbf{W}^{G}_{i,j}}{s_{i,j}}\right\rceil,\,-1,\,1\right)\ 
  \text{with}\ s_{i,j} = \max\!\left(\dfrac{\beta}{G}\displaystyle\sum_{k}|W^{G}_{i,j,k}|,\ \epsilon\right),
  & \text{if}\ n = 1.58 \\[6pt]
  \left\lfloor\dfrac{\mathbf{W}^{G}_{i,j}}{s_{i,j}}\right\rceil\ 
  \text{with}\ s_{i,j} = \max\!\left(\dfrac{\displaystyle\max_k|W^{G}_{i,j,k}|}{2^{n-1}-1},\ \epsilon\right),
  & \text{if}\ n \in \{4,\,8\}
\end{cases}
\end{equation}
% - $\beta$ is the hyperparameter for adjusting the quantization threshold
% - $\epsilon$ 为防止除零的极小值
%
where $\lfloor \cdot \rceil$ denotes rounding to the nearest integer, $s_{i,j}$ is the scaling factor, $\beta$ is a tunable scaling coefficient for ternarization, and $\epsilon$ is a small constant that prevents division by zero.
The 1.58-bit branch, as $\log_2 3 \approx 1.58$, quantizes each weight to the ternary set
$\{-1,0,+1\}$, with the scaling factor derived from the group mean absolute value.
The $n$-bit branch, where $n \in \{4,8\}$, quantizes weights to the symmetric integer range such as $[-7,\,7]$ for 4-bit and $[-127,\,127]$ for 8-bit, with the scaling factor determined by the group-wise maximum absolute value.
Specifically, the activation groups are uniformly quantized to 8-bit under weight-activation quantization or retained at 16-bit under weight-only quantization.

\subsection{Strategies of structural quantization with mixed precision}
\label{apx:mpqad-allocation}
% 图 super-group

\begin{figure*}[!ht]
    \centering
    \includegraphics[width=0.80\textwidth ]{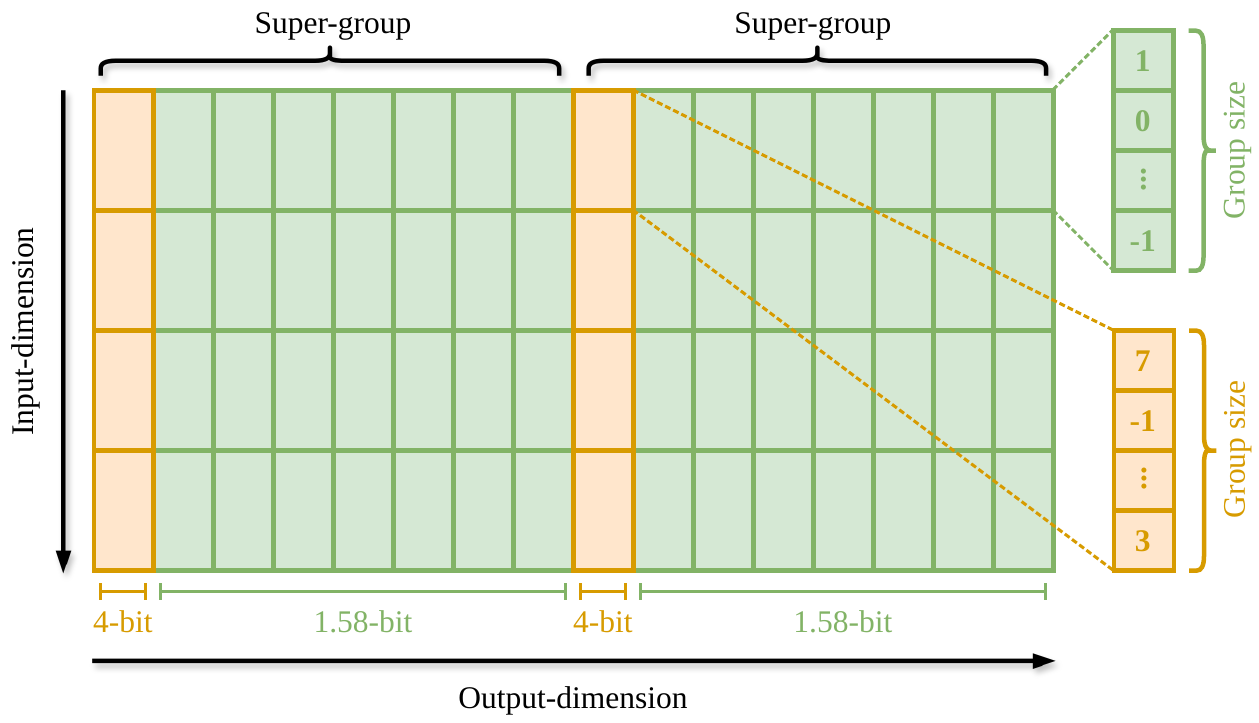}
    \caption{Super-group allocation for weight matrices, visualized via the transposed matrix $\mathbf{W}^T$.}
    \label{fig:super-group-allocation}
\end{figure*}

\begin{figure*}[!ht]
    \centering
    \includegraphics[width=0.80\textwidth ]{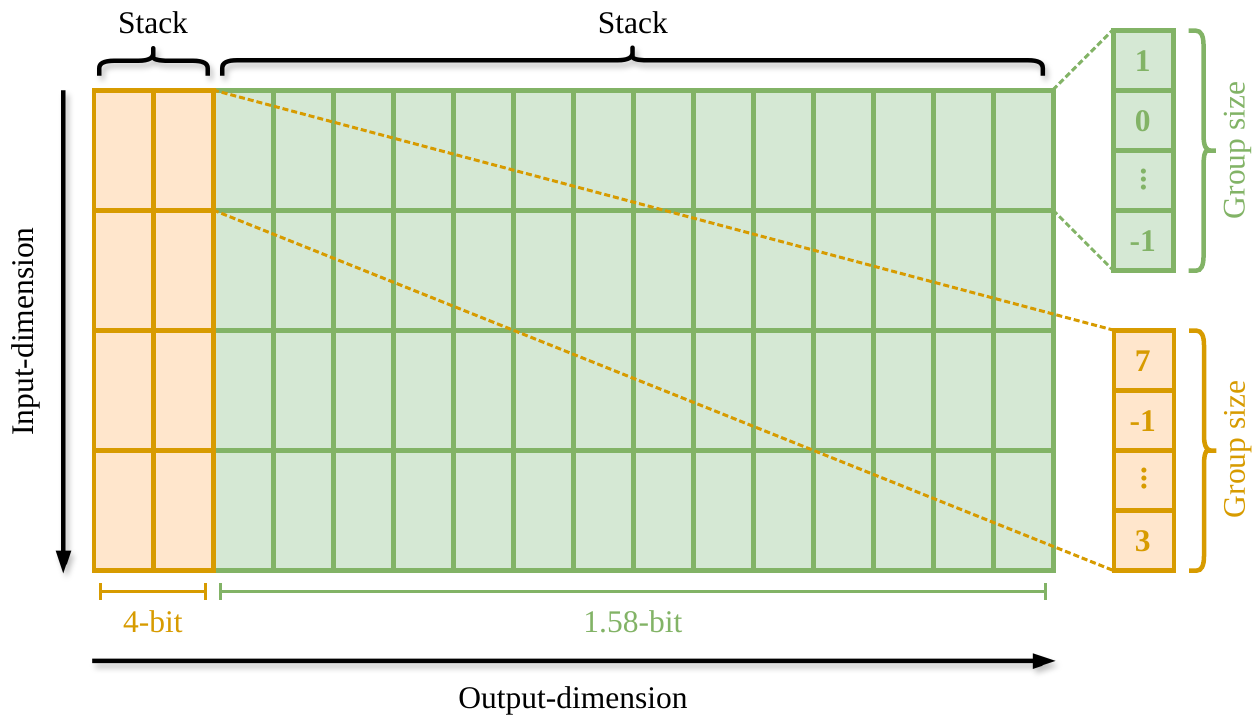}
    \caption{Stacked allocation for weight matrices, visualized via the transposed matrix $\mathbf{W}^T$.}
    \label{fig:stacked-allocation}
\end{figure*}

In this section, we provide strategies on how to allocate high-precision and low-precision groups within weight matrices in our proposed Structural Quantization with Mixed Precision (SQMP) module. Figure~\ref{fig:stacked-allocation} shows the stacked allocation, while Figure~\ref{fig:super-group-allocation} depicts the super-group allocation. For instance, setting $\rho=1/8$ places one 4-bit output channel followed by seven 1.58-bit output channels within each super-group, culminating in an effective bit-width of roughly 1.88-bit.

\subsection{Proof of super-group allocation in SQMP}
\label{apx:optimal-allocation-proof}

In this section, we justify the super-group allocation by showing that, under a bounded-variation model of salience drift during training, it minimizes a discrepancy-based worst-case upper bound on the salience-alignment error among the allocation schemes considered in this paper.

% --------------------------------------------------------------------
\paragraph{Definition.}
Let $\mathbf{W}\in\mathbb{R}^{d_{\mathrm{out}}\times d_{\mathrm{in}}}$ and $\mathbf{X}\in\mathbb{R}^{d_{\mathrm{in}}\times L}$ denote the full-precision weights and activations, and let
\begin{equation}
    \Delta\mathbf{Y}\;=\;\mathbf{W}\mathbf{X}-Q(\mathbf{W})\mathbf{X}
\end{equation}
denote the quantization error. For each output row $i$, let $a_i\in\{0,1\}$ indicate the precision assignment, where $a_i=1$ denotes a 4-bit assignment and $a_i=0$ denotes a 1.58-bit assignment. Under the bit-width budget and specific proportion,
\begin{equation}
    \frac{1}{d_{\mathrm{out}}}\sum_{i=1}^{d_{\mathrm{out}}} a_i\;=\;\rho,
    \ \ \text{and} \ \ 
    N\;:=\;\sum_{i=1}^{d_{\mathrm{out}}}a_i\;=\;\rho\,d_{\mathrm{out}},
\end{equation}
we adopt the row-separable surrogate
\begin{equation}
\label{eq:quant-error-bound-rigorous}
    \mathcal{L}(\mathbf{a})
    \;\lesssim\;
    \sum_{i=1}^{d_{\mathrm{out}}}
    \bigl((1-a_i)\,e_L+a_i\,e_H\bigr)\,S_i
    \;=\;
    \sum_{i=1}^{d_{\mathrm{out}}}
    \bigl(e_L-a_i(e_L-e_H)\bigr)\,S_i,
\end{equation}
where $S_i\ge 0$ is the local salience of the $i$-th row and $e_H<e_L$ are precision-dependent errors corresponding to 4-bit and 1.58-bit quantization, respectively. Since $e_L$ and $e_H$ are constants and $\sum_i S_i$ is independent of $\mathbf{a}$, minimizing \eqref{eq:quant-error-bound-rigorous} is equivalent to maximizing the salience alignment
\begin{equation}
\label{eq:alignment-def}
    \mathcal{A}(\mathbf{a})\;:=\;\sum_{i=1}^{d_{\mathrm{out}}} a_i\,S_i.
\end{equation}

% --------------------------------------------------------------------
\paragraph{Assumption.}
In Post-Training Quantization (PTQ), the salience profile $S_i$ is static, which enables direct salience-driven assignments~\citep{lin2024awq}. Conversely, Quantization-Aware Training (QAT) and Quantization-Aware Distillation (QAD) continuously update the model through data-driven gradients, and activation outliers are empirically observed to shift across channels during training~\citep{heo2024adadim}, causing row salience $S_i$ to fluctuate accordingly. However, as the parameters and activations are tightly bounded by regularization mechanisms such as weight decay and normalization layers, we assume that the continuous extension $S:[0,1]\!\to\!\mathbb{R}_{\ge 0}$ of the salience profile is a function of bounded total variation, $V(S)<\infty$.

\paragraph{Worst-case Formulation.}
Since the exact trajectory of $S$ is unpredictable \textit{a priori} during training, an allocation strategy that is robust to such salience drift should minimize the worst-case alignment error over all functions $S$ satisfying the bounded total variation assumption. Formally, we identify each row index $i\in\{1,\dots,d_{\mathrm{out}}\}$ with the midpoint $x_i=(i-\tfrac12)/d_{\mathrm{out}}\in[0,1]$ and write $S_i=S(x_i)$. Let $P=\{p_1,\dots,p_N\}\subset[0,1]$ denote the normalized locations of the $N$ high-precision rows. Then
\begin{equation}
    \mathcal{A}(P)\;=\;\sum_{k=1}^{N} S(p_k),
\end{equation}
and maximizing $\mathcal{A}(P)$ robustly under the unknown but drift-bounded $S$ reduces to requiring the empirical average $\frac{1}{N}\sum_{k} S(p_k)$ to uniformly approximate the ideal mean $\int_0^1 S(x)\,dx$. By the one-dimensional Koksma--Hlawka inequality,
\begin{equation}
\label{eq:koksma-hlawka}
    \left|
    \frac{1}{N}\sum_{k=1}^{N}S(p_k)-\int_0^1 S(x)\,dx
    \right|
    \;\le\;
    V(S)\,D_N^*(P),
\end{equation}
where the discrepancy of $P$ is defined as
\begin{equation}
    D_N^*(P)\;:=\;
    \sup_{t\in[0,1]}
    \left|
    \frac{1}{N}\sum_{k=1}^{N}\mathbf{1}\{p_k\le t\}-t
    \right|.
\end{equation}
Since the bound \eqref{eq:koksma-hlawka} is uniform over all bounded-variation $S$, minimizing $D_N^*(P)$ tightens the worst-case upper bound on the salience-alignment error.

% --------------------------------------------------------------------
\paragraph{Proposition.}
Among the random, stacked, and super-group allocations, the super-group allocation attains the smallest order of discrepancy, namely $\Theta(N^{-1})$. Consequently, under the surrogate \eqref{eq:quant-error-bound-rigorous}, it minimizes the Koksma--Hlawka worst-case upper bound \eqref{eq:koksma-hlawka} up to an absolute constant.

% --------------------------------------------------------------------
\paragraph{Proof.}
We compute the discrepancy for each of the three allocations.

\textbf{1. Random allocation.} The $N$ high-precision rows are distributed uniformly at random, yielding
$p_k\overset{\mathrm{i.i.d.}}{\sim}\mathrm{Unif}[0,1]$.
Standard empirical process bounds imply that the discrepancy satisfies
\begin{equation}
    D_N^*(P_{\mathrm{rand}})
    =
    \mathcal{O}_p(N^{-1/2}) \ .
\end{equation}

\textbf{2. Stacked allocation.} All $N$ high-precision rows are clustered contiguously at one end of the output dimension, yielding
\begin{equation}
    P_{\mathrm{stack}}
    \;=\;
    \left\{
    \frac{0.5}{d_{\mathrm{out}}},
    \frac{1.5}{d_{\mathrm{out}}},
    \dots,
    \frac{N-0.5}{d_{\mathrm{out}}}
    \right\} \ .
\end{equation}
Since all points lie in a sub-interval of length $\rho$, taking $t=\rho$ in the definition of $D_N^*$ gives a deviation of $1-\rho$. Thus, the discrepancy is constant
\begin{equation}
    D_N^*(P_{\mathrm{stack}})
    \;=\;
    1-\rho
    \;=\;
    \Theta(1) \ .
\end{equation}

\textbf{3. Super-group allocation (ours).} The 4-bit rows are placed on a deterministic equidistant grid with period $\lfloor 1/\rho\rceil$ along the output dimension. Then, the normalized pattern is the midpoint grid
\begin{equation}
    P_{\mathrm{super}}
    \;=\;
    \left\{
    \frac{2k-1}{2N}
    \right\}_{k=1}^{N} \ .
\end{equation}
For any $t\in[0,1]$, the number of points in $[0,t]$ is $\lfloor Nt+\tfrac12\rfloor$, so that
\begin{equation}
    \left|
    \frac{1}{N}\sum_{k=1}^{N}\mathbf{1}\{p_k\le t\}-t
    \right|
    \;\le\;
    \frac{1}{2N} \ .
\end{equation}
While rounding row indices to discrete integers introduces a minor mapping perturbation bounded by $\mathcal{O}(1/d_{\mathrm{out}})$, since $d_{\mathrm{out}} = N/\rho$, this shift is $\mathcal{O}(1/N)$. Thus, the super-group allocation rigorously maintains the optimal discrepancy order,
\begin{equation}
    D_N^*(P_{\mathrm{super}})
    \;=\;
    \Theta(N^{-1}) \ .
\end{equation}

By classical one-dimensional discrepancy theory, every $N$-point set $P\subset[0,1]$ satisfies $D_N^*(P)\ge c/N$ for some absolute constant $c>0$. Hence, the super-group allocation already attains the optimal order of one-dimensional discrepancy. Combining this with the Koksma--Hlawka inequality \eqref{eq:koksma-hlawka} proves the stated worst-case optimality.

\subsection{Analysis and discussion}

This periodic super-group allocation addresses a limitation of static salience-based assignments in QAT and QAD.
By mathematically minimizing the discrepancy $D_N^*(P)$ to its theoretical lower bound, our structural mixed-precision layout minimizes the discrepancy-based worst-case upper bound under arbitrary and unpredictable salience shifts $V(S)$. It ensures that each token accumulates exactly $\rho$ of 4-bit contributions, thereby effectively decoupling model performance from unpredictable fluctuations in salience. Furthermore, the interleaved super-groups inherently serve as evenly spaced, high-precision buffers along the output dimension. This uniform distribution of precision guarantees that no contiguous block of output features suffers from concentrated degradation, effectively mitigating clustered quantization errors and yielding a smaller discrepancy bound than either the stacked or the random allocation.
Beyond algorithmic stability, the random allocation is not hardware-friendly, while our deterministic, repeating structure aligns with hardware execution granularities, supporting coalesced memory access and potentially improving throughput in low-level kernel implementations.

%%%%%%%%%%%%%%%%%%%%%%%%%%%%%%%%%%%%%%%%%%%%%%%%%%%%%%%%%%%%

\section{Pattern of important features}
\label{apx:pattern-of-features}

\begin{figure*}[!ht]
    \centering
    \includegraphics[width=0.8\textwidth ]{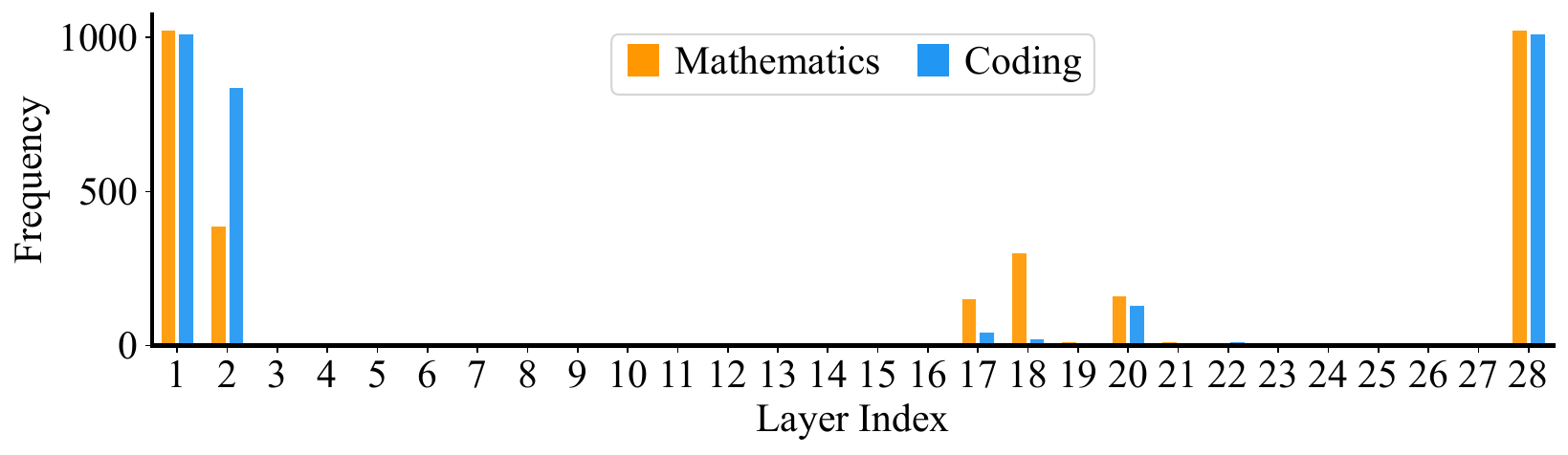}
    \caption{Frequency of the $k$ layers with the lowest $c_l$ ($k=3$).}
    \label{fig:top3-frequency}
\end{figure*}

\begin{figure*}[!ht]
    \centering
    \includegraphics[width=0.8\textwidth ]{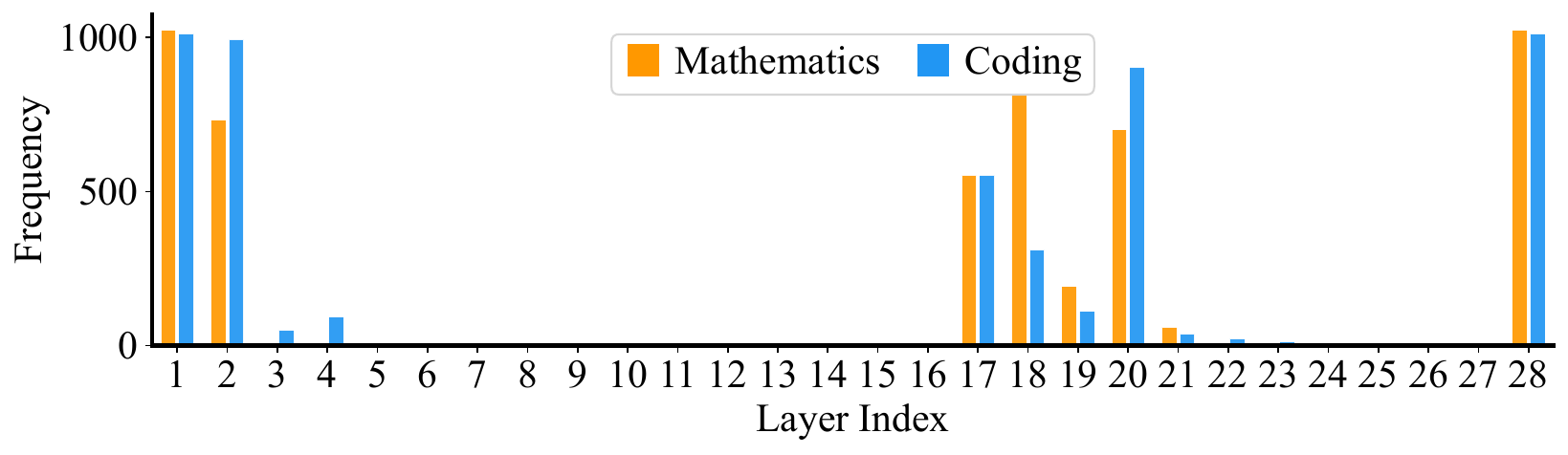}
    \caption{Frequency of the $k$ layers with the lowest $c_l$ ($k=5$).}
    \label{fig:top5-frequency}
\end{figure*}

\begin{figure*}[!ht]
    \centering
    \includegraphics[width=0.8\textwidth ]{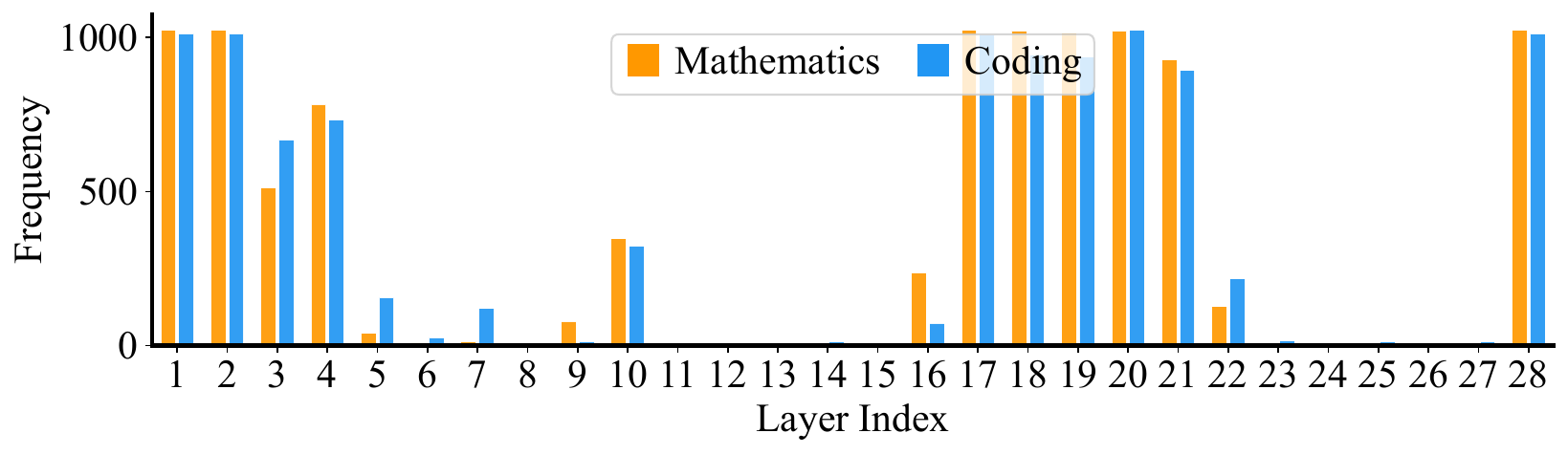}
    \caption{Frequency of the $k$ layers with the lowest $c_l$ ($k=10$).}
    \label{fig:top10-frequency}
\end{figure*}

\begin{figure*}[!ht]
    \centering
    \includegraphics[width=\textwidth ]{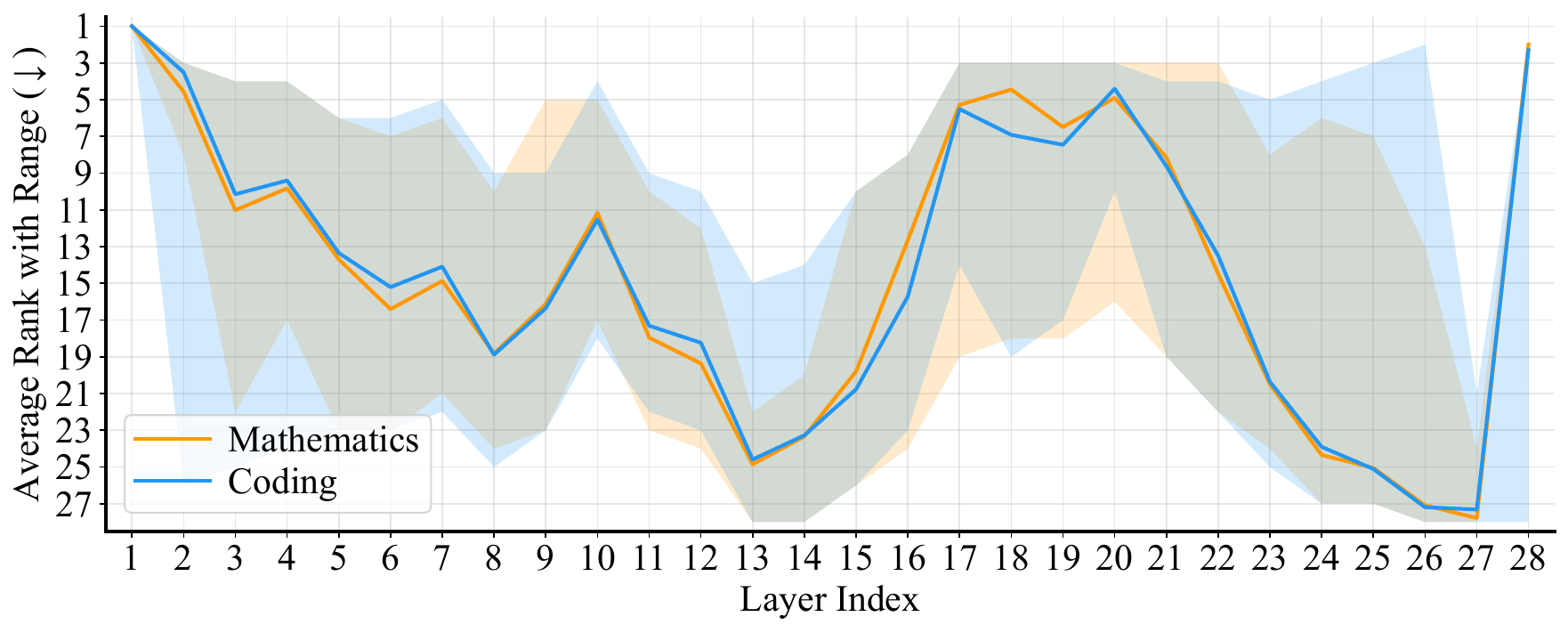}
    \caption{Average rank of each layer along with its corresponding range.}
    \label{fig:average-rank}
\end{figure*}

In Section~\ref{subsec:LAFD}, we adopt cosine similarity as an importance metric to quantify transformations across feature layers. In this section, we investigate how important feature patterns vary across data domains, thereby demonstrating the limitations of pre-specified layer supervision. We analyze two distinct domains, mathematics and coding, from our training corpus. The mathematical dataset is \href{https://hf-mirror.com/datasets/TIGER-Lab/MathInstruct}{TIGER-Lab/MathInstruct}~\citep{yue2023mammoth} and the coding dataset is \href{https://huggingface.co/datasets/jtatman/python-code-dataset-500k}{jtatman/python-code-dataset-500k}. Specifically, we randomly sample 1024 instances from each domain, using seed 42, and process them with the full-precision Qwen3-1.7B to calculate the similarities between consecutive layers. We aggregate the frequency at which each layer ranks among the $k$ layers with the lowest similarity in Figures~\ref{fig:top3-frequency}, \ref{fig:top5-frequency}, and \ref{fig:top10-frequency}. We further depict the average rank of each layer, along with its corresponding variance, in Figure~\ref{fig:average-rank}.

There are three key observations:
(1) In Figure~\ref{fig:top3-frequency} ($k=3$), frequency distributions diverge significantly: mathematical data heavily transforms layers 17 and 18, which coding data rarely alters, whereas coding data modifies layer 2 more frequently;
(2) In Figures~\ref{fig:top5-frequency} and \ref{fig:top10-frequency} ($k=5, 10$), this structural divergence persists, notably at $k=5$ where mathematical data transforms layer 18 over 750 times compared to only ${\sim}300$ times for coding data;
(3) In Figure~\ref{fig:average-rank}, the average rank analysis further confirms this discrepancy, revealing that mathematical data prominently transforms layers 17 through 22, while coding data primarily targets layers 17 through 20 and uniquely emphasizes layer 25.

Observations (1)--(2) indicate that the frequency of largely transformed intermediate layers is domain-dependent. Observation (3) reveals that critical representational bottlenecks are specialized and only partially overlap across domains.
Observations (1)--(3) demonstrate that statically encompassing all these disjoint critical layers requires a large union, which induces redundant and over-regularized computations. This additionally motivates the LAFD configuration $k=3$ for our training hyperparameters in Section~\ref{sec:experimental-configurations}, where dynamically selecting 3 layers strictly concentrates the distillation signal on the most essential representational bottlenecks.

These results reveal that the pattern of important intermediate layers is domain-dependent, rendering fixed-layer feature distillation inefficient for heterogeneous datasets. Motivated by this structural uncertainty, we propose Adaptive Feature Distillation in Section~\ref{subsec:LAFD}.

% 如果说 top-3 就选 6 层来训练，那么每次 fd 的时候，都只有3层是有用的，另外3层的计算都是redundant的

% How to configure K_EAKLD => stat entropy of response tokens
\section{Mixing coefficient of forward-reverse KL divergence}
\label{apx:teacher-confidence}

\begin{table}[!ht]
  \centering
  \caption{The proportions of mismatched tokens among high-confidence tokens.}
  \label{tab:mismatch}
  \setlength{\tabcolsep}{3pt}
  \begin{tabular}{l c c >{\columncolor{gray!20}\bfseries}c c >{\columncolor{gray!20}}c}
    \toprule
    \multirow{2}{*}{\textbf{Data types}} & \multirow{2}{*}{\textbf{Thresholds}} & \multicolumn{2}{c}{\textbf{Human-annotated data}} & \multicolumn{2}{c}{\textbf{Distilled data}} \\
    \cmidrule(lr){3-4} \cmidrule(lr){5-6}
    & & \textbf{High-conf (\%)} & \multicolumn{1}{c}{\textbf{Mismatched (\%)}} & \textbf{High-conf (\%)} & \multicolumn{1}{c}{\textbf{Mismatched (\%)}} \\
    \midrule
    \multirow{5}{*}{Mathematics} & 0.60 & 79.18 & 16.78 & 86.16 & 2.98 \\
    & 0.70 & 72.36 & 13.64 & 81.02 & 1.76 \\
    & 0.80 & 65.25 & 10.54 & 75.61 & 1.14 \\
    & 0.90 & 56.48 & 7.28 & 68.61 & 0.74 \\
    & 0.95 & 49.47 & 5.36 & 62.83 & 0.60 \\
    \midrule
    \multirow{5}{*}{Coding} & 0.60 & 82.96 & 11.11 & 83.14 & 2.46 \\
    & 0.70 & 76.67 & 8.51 & 77.38 & 1.31 \\
    & 0.80 & 70.15 & 6.13 & 71.29 & 0.86 \\
    & 0.90 & 61.77 & 3.77 & 63.60 & 0.63 \\
    & 0.95 & 55.07 & 2.41 & 57.34 & 0.56 \\
    \bottomrule
  \end{tabular}
\end{table}

\begin{figure*}[!ht]
    \centering
    \includegraphics[width=0.75\textwidth ]{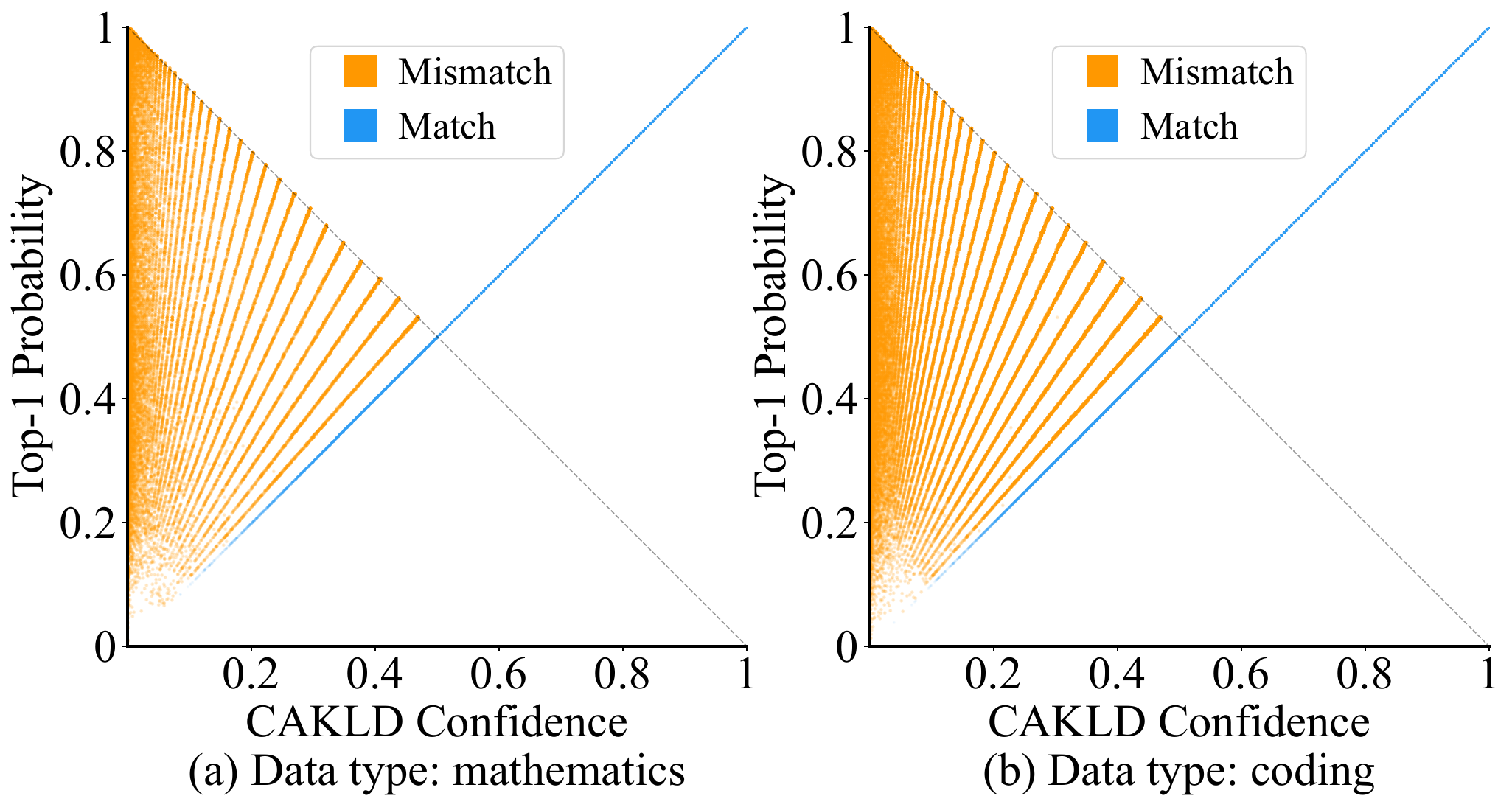}
    \caption{Scatter plots of top-1 probability versus CAKLD confidence on human-annotated datasets. Mismatched tokens frequently exhibit high top-1 probabilities despite near-zero CAKLD confidence.}
    \label{fig:cakld-confidence}
\end{figure*}

The CAKLD method introduced in BitDistiller~\citep{du2024bitdistiller} employs the label token probability as the mixing coefficient to balance the forward and reverse KL divergences. This approach relies entirely on internally distilled datasets generated by the teacher model. Consequently, it struggles to leverage diverse corpora, such as high-quality human-annotated datasets or synthetic data generated by leading external models.
Using the same datasets in Appendix~\ref{apx:pattern-of-features}, we randomly select 1024 samples from the mathematical and coding datasets, using seed 42. These datasets comprise both human-annotated and synthetic data. We prompt the full-precision Qwen3-0.6B to generate distilled responses using a maximum of 1024 tokens, a temperature of 0.7, a top-p of 0.8, and a top-k of 20. Subsequently, we collect the teacher output distributions via forward passes. CAKLD confidence denotes the probability of the label token. We define high-confidence tokens as those for which the teacher assigns a probability exceeding a specified threshold. Mismatch tokens are instances where the teacher's top-1 predicted token differs from the provided label. Table~\ref{tab:mismatch} presents the mismatch proportion among the high-confidence tokens, and Figure~\ref{fig:cakld-confidence} exhibits all mismatched tokens.

There are three observations:
(1) In Table~\ref{tab:mismatch}, for mathematical data, the proportion of mismatched tokens within high-confidence tokens is significantly larger on the human-annotated and synthetic data compared to internally distilled data across all thresholds, scaling from 16.78\% versus 2.98\% at a 0.60 threshold down to 5.36\% versus 0.60\% at a 0.95 threshold;
(2) In Table~\ref{tab:mismatch}, coding data exhibits an identical trend, with mismatched proportions on human-annotated and synthetic data reaching 11.11\% compared to 2.46\% on distilled data at the 0.60 threshold;
(3) In Figure~\ref{fig:cakld-confidence}, both mathematical and coding data reveal dense mismatch distributions where the teacher exhibits high top-1 probability despite the corresponding CAKLD confidence approaching zero.

Observations (1)--(3) demonstrate that CAKLD yields substantial mismatched errors when applied to human-annotated and synthetic data. This strict dataset dependency prevents quantized models from leveraging optimal and heterogeneous training recipes. Motivated by this limitation, we propose Entropy-Aware KL Divergence in Section~\ref{subsec:EAKLD} to generalize logit distillation across various datasets.

%%%%%%%%%%%%%%%%%%%%%%%%%%%%%%%%%%%%%%%%%%%%%%%%%%%%%%%%%%%%

\section{Additional details of experiments}

\subsection{Training data}
\label{apx:training-data} % Due to space limitations

\begin{table}[!ht]
\caption{Details of datasets used for training.}
\label{tab:datasets}
\centering
\small
\setlength{\tabcolsep}{6pt}
\begin{tabular*}{\textwidth}{@{\extracolsep{\fill}}lllll}
\toprule
\textbf{\#} & \textbf{Datasets} & \textbf{Subsets} & \textbf{Split} & \textbf{Data sizes} \\
\midrule
1 & BAAI/Infinity-Instruct~\citep{li2025infinity} & 7M\_domains & train & 7.45M \\
2 & BAAI/Infinity-Instruct & Gen & train & 1.46M \\
3 & \href{https://huggingface.co/datasets/allenai/tulu-v3.1-mix-preview-4096-OLMoE}{allenai/tulu-v3.1-mix-preview-4096-OLMoE} & -- & train & 0.61M \\
4 & a-m-team/AM-DeepSeek-R1-Distilled-1.4M~\citep{zhao2025amdistill} & am\_0.5M+am\_0.9M & train & 1.40M \\
5 & Mixed Downstream Datasets~\citep{bisk2020piqa,clark2019boolq,clark2018arc,lin2022truthfulqa,mihaylov2018openbookqa,sakaguchi2021winogrande,sap2019socialiqa,zellers2019hellaswag} & -- & train & 0.24M \\
6 & BAAI/Infinity-Instruct & 7M\_core & train & 1.48M \\
7 & HuggingFaceM4/TGIF~\citep{li2016tgif} & -- & train & 10K \\
\bottomrule
\end{tabular*}
\end{table}

% In Table~\ref{tab:datasets}, we provide details of training datasets, including their sources, subsets, splits, and sizes. Since we adopt the non-reasoning mode for Qwen3 models, the distilled 1.4 million DeepSeek samples are used after removing chain-of-thought traces. Datasets 1--5 are used to train Qwen3 models; Datasets 5 and 6 are used to train MobieLLM-350M, and Dataset 7 is used to train Qwen2.5-Omni-7B.

Table~\ref{tab:datasets} summarizes the sources, subsets, and sizes of our training corpora. We construct tailored data mixtures for each model architecture. We train the Qwen3 models on datasets 1 through 5. Because we configure Qwen3 in a non-reasoning mode, we strip all chain-of-thought traces from the distilled DeepSeek samples prior to training. We train MobileLLM-350M using datasets 5 and 6. Finally, we train the multimodal Qwen2.5-Omni-7B model on dataset 7. Additionally, sequences are truncated to a maximum length of 1024 tokens for all datasets during training.

\subsection{Evaluation protocols}
\label{apx:eval-protocols}

In Table~\ref{tab:benchmarks}, we summarize the evaluation protocols, including downstream tasks with their categories, N-shot settings, scoring methods, and metrics. 0-shot* denotes the 0-shot evaluation format, while the models are trained to training splits from these domains. In Appendix~\ref{apx:generalization-held-out}, we provide results validating robust generalization on held-out benchmarks.

\begin{table}[!ht]
\caption{Overview of evaluation benchmarks used for \textsc{EdgeRazor}.}
\label{tab:benchmarks}
\centering
\small
\setlength{\tabcolsep}{10pt}
\begin{tabular*}{\textwidth}{@{\extracolsep{\fill}}lllll}
\toprule
\textbf{Categories} & \textbf{Tasks} & \textbf{N-shot} & \textbf{Scoring methods} & \textbf{Metrics} \\
\midrule
\multirow{7}{*}{Commonsense reasoning}
 & ARC-e~\citep{clark2018arc}       & 0-shot* & Log-likelihood & Acc\_norm \\
 & ARC-c~\citep{clark2018arc}       & 0-shot* & Log-likelihood & Acc\_norm \\
 & HellaSwag~\citep{zellers2019hellaswag}   & 0-shot* & Log-likelihood & Acc\_norm \\
 & BoolQ~\citep{clark2019boolq}       & 0-shot* & Log-likelihood & Acc \\
 & PIQA~\citep{bisk2020piqa}        & 0-shot* & Log-likelihood & Acc\_norm \\
 & Winogrande~\citep{sakaguchi2021winogrande}  & 0-shot* & Log-likelihood & Acc \\
 & SIQA~\citep{sap2019socialiqa}        & 0-shot* & Log-likelihood & Acc \\
\midrule
Reading comprehension & OpenBookQA~\citep{mihaylov2018openbookqa}  & 0-shot* & Log-likelihood & Acc\_norm \\
\midrule
Trustworthiness & Ethics~\citep{hendrycks2021ethics}      & 0-shot* & Log-likelihood & Acc \\
\midrule
Truthfulness & TruthfulQA2~\citep{lin2022truthfulqa} & 0-shot & Log-likelihood & Acc \\
\midrule
Knowledge      & MMLU~\citep{hendrycks2021mmlu}          & 0-shot & Log-likelihood & Acc \\
\midrule
Instruction following & IF-Eval~\citep{zhou2023ifeval}   & 0-shot & Generation     & Prompt Strict Acc \\
\midrule
Mathematics           & GSM8K~\citep{cobbe2021gsm8k}     & 5-shot & Generation & Acc \\
\midrule
Coding         & HumanEval~\citep{chen2021humaneval} & 0-shot & Generation     & Pass@1 \\
\midrule
\multirow{2}{*}{Video understanding}
 & Video-MME~\citep{fu2024videomme}   & 0-shot & Generation     & Acc \\
 & MLVU~\citep{zhou2024mlvu}        & 0-shot & Generation     & Acc \\
\bottomrule
\end{tabular*}
\end{table}

\subsection{Per-task evaluation results}
\label{apx:effectiveness} % Due to space limitations

In this section, we provide the complete per-task results across comprehensive downstream tasks. The models include both base and instruction-tuned LLMs.
We compare against a comprehensive suite of recent PTQ and QAT baselines, including GPTQ~\citep{frantar2023gptq}, OmniQuant~\citep{shao2024omniquant}, AWQ~\citep{lin2024awq}, AQLM~\citep{egiazarian2024aqlm}, BiLLM~\citep{huang2024billm}, QuIP\#~\citep{tseng2024quipsharp}, AutoRound~\citep{cheng2024autoround}, VPTQ~\citep{liu2024vptq}, QTIP~\citep{tseng2024qtip}, ARB-LLM~\citep{li2025arbllm}, GPTAQ~\citep{li2025gptaq}, Slim-LLM+~\citep{huang2025slimllm}, Q-Palette~\citep{lee2025qpalette}, LQER~\citep{zhang2024lqer}, QuaRot~\citep{ashkboos2024quarot}, ABQ-LLM~\citep{li2025abq}, SpinQuant~\citep{liu2025spinquant}, QoQ~\citep{lin2025qserve}, FlatQuant~\citep{sun2025flatquant}, EfficientQAT~\citep{chen2024efficientqat}, and ParetoQ~\citep{liu2025paretoq}.
For the baselines, the group size is set to 64 for MobileLLM-350M and 128 for the Qwen models.
Tables~\ref{tab:detailed-weight-only-quant-on-mobilellm350m} and \ref{tab:detailed-wakv-quant-on-mobilellm350m} present per-task performance for MobileLLM-350M.
Tables~\ref{tab:detailed-weight-only-quant-on-qwen3-0.6b} and \ref{tab:detailed-wakv-quant-on-qwen3-0.6b} list per-task performance for Qwen3-0.6B.
Tables~\ref{tab:detailed-weight-only-quant-on-qwen3-1.7b} and \ref{tab:detailed-weight-activation-quant-on-qwen3-1.7b} provide per-task performance for Qwen3-1.7B. Notably, the best performance is indicated in \textbf{bold}, and the second-best is \underline{underlined}. Scores of 0.00 for full-precision models on tasks such as GSM8K and HumanEval indicate a lack of capability, whereas scores of 0.00 for quantized models on these tasks demonstrate severe capability degradation.

\begin{table}[!ht]
  \centering
  \caption{Detailed performance of weight-only quantization methods on MobileLLM-350M.}
  \label{tab:detailed-weight-only-quant-on-mobilellm350m}
  \setlength{\tabcolsep}{2pt}
  \resizebox{\textwidth}{!}{%
\begin{tabular}{lcccccccccccccccc}
    \toprule
    \textbf{Methods} & \textbf{W-A-KV} & \textbf{ARC-e} & \textbf{ARC-c} & \textbf{HellaS.} & \textbf{BoolQ} & \textbf{PIQA} & \textbf{WinoG.} & \textbf{SIQA} & \textbf{OBQA} & \textbf{Ethics} & \textbf{Tr.QA2} & \textbf{MMLU} & \textbf{GSM8K} & \textbf{HumanE.} & \textbf{Average} \\
    \midrule
    \textbf{BF16}          & 16-16-16    & 64.94 & 35.49 & 52.87 & 58.96 & 70.84 & 56.35 & 40.79 & 40.20 & 53.98 & 37.44 & 23.52 & 0.00 & 0.00 & 41.18 \\
    \midrule
    \textbf{OmniQuant}     & 4-16-16     & 63.30 & 34.64 & 51.79 & 58.62 & 70.62 & 54.46 & 40.12 & 38.40 & 48.76 & 37.48 & 24.26 & 0.00 & 0.00 & 40.19 \\
    \textbf{OmniQuant}     & 3-16-16     & 61.07 & 33.53 & 48.73 & 60.58 & 68.28 & 54.78 & 39.61 & 36.00 & 51.56 & 39.90 & 23.14 & 0.00 & 0.00 & 39.78 \\
    \textbf{OmniQuant}     & 2-16-16     & 43.98 & 24.57 & 33.42 & 56.48 & 58.11 & 49.41 & 35.67 & 29.60 & 45.26 & 43.93 & 23.18 & 0.00 & 0.00 & 34.12 \\
    \midrule
    \textbf{AWQ}           & 4-16-16     & 63.80 & 35.49 & 52.10 & 59.05 & 70.24 & 54.38 & 39.97 & 39.20 & 54.12 & 37.85 & 24.16 & 0.00 & 0.00 & 40.80 \\
    \textbf{AWQ}           & 3-16-16     & 62.67 & 33.62 & 50.26 & 61.01 & 68.61 & 54.54 & 39.76 & 37.40 & 46.94 & 39.91 & 24.23 & 0.00 & 0.00 & 39.92 \\
    \textbf{AWQ}           & 2-16-16     & 36.15 & 25.00 & 30.92 & 48.65 & 56.20 & 48.93 & 34.60 & 27.40 & 47.63 & 51.23 & 23.23 & 0.00 & 0.00 & 33.07 \\
\bottomrule
  \end{tabular}%
  }
\end{table}

\begin{table}[!ht]
  \centering
  \caption{Detailed performance of weight-activation quantization methods on MobileLLM-350M.}
  \label{tab:detailed-wakv-quant-on-mobilellm350m}
  \setlength{\tabcolsep}{2pt}
  \resizebox{\textwidth}{!}{%
  \begin{tabular}{lcccccccccccccccc}
    \toprule
    \textbf{Methods} & \textbf{W-A-KV} & \textbf{ARC-e} & \textbf{ARC-c} & \textbf{HellaS.} & \textbf{BoolQ} & \textbf{PIQA} & \textbf{WinoG.} & \textbf{SIQA} & \textbf{OBQA} & \textbf{Ethics} & \textbf{Tr.QA2} & \textbf{MMLU} & \textbf{GSM8K} & \textbf{HumanE.} & \textbf{Average} \\
    \midrule
    \textbf{BF16} & 16-16-16 & 64.94 & 35.49 & 52.87 & 58.96 & 70.84 & 56.35 & 40.79 & 40.20 & 53.98 & 37.44 & 23.52 & 0.00 & 0.00 & 41.18 \\
    \midrule
    \textbf{OmniQuant}      & 4-8-8    & 63.93 & 34.81 & 51.86 & 57.80 & 70.18 & 55.01 & 40.17 & 37.80 & 50.93 & 37.38 & 24.37 & 0.00 & 0.00 & 40.33 \\
    \textbf{OmniQuant}      & 3-8-8    & 61.41 & 32.08 & 48.78 & 60.21 & 68.28 & 54.14 & 39.56 & 35.80 & 52.66 & 38.30 & 23.57 & 0.00 & 0.00 & \underline{39.60} \\
    \textbf{OmniQuant}      & 2-8-8    & 44.49 & 23.21 & 33.21 & 62.02 & 57.13 & 50.99 & 34.95 & 29.00 & 46.75 & 44.85 & 23.42 & 0.00 & 0.00 & 34.62 \\
    \midrule
    \textbf{LQER}           & 4-8-8   & 64.02 & 36.01 & 51.84 & 56.61 & 70.78 & 53.91 & 40.02 & 39.40 & 53.17 & 38.36 & 24.08 & 0.00 & 0.00 & \underline{40.63} \\
    \textbf{LQER}           & 3-8-8   & 58.54 & 32.17 & 48.82 & 58.26 & 67.46 & 53.51 & 38.84 & 37.20 & 49.53 & 41.88 & 24.63 & 0.00 & 0.00 & 39.30 \\
    \textbf{LQER}           & 2-8-8   & 35.86 & 23.29 & 29.21 & 44.43 & 54.90 & 49.17 & 35.52 & 25.40 & 53.12 & 45.15 & 23.31 & 0.00 & 0.00 & 32.26 \\
\bottomrule
  \end{tabular}%
  }
\end{table}

\begin{table}[!ht]
  \centering
  \caption{Detailed performance of weight-only quantization methods on Qwen3-0.6B.}
  \label{tab:detailed-weight-only-quant-on-qwen3-0.6b}
  \setlength{\tabcolsep}{2pt}
  \resizebox{\textwidth}{!}{%
  \begin{tabular}{lccccccccccccccccc}
    \toprule
    \textbf{Methods} & \textbf{W-A-KV} & \textbf{ARC-e} & \textbf{ARC-c} & \textbf{HellaS.} & \textbf{BoolQ} & \textbf{PIQA} & \textbf{WinoG.} & \textbf{SIQA} & \textbf{OBQA} & \textbf{Ethics} & \textbf{Tr.QA2} & \textbf{MMLU} & \textbf{IFEval} & \textbf{GSM8K} & \textbf{HumanE.} & \textbf{Average} \\
    \midrule
    \textbf{BF16}         & 16-16-16 & 56.02 & 34.04 & 47.23 & 64.04 & 67.36 & 56.04 & 39.20 & 31.20 & 47.70 & 42.84 & 40.12 & 58.41 & 41.54 & 37.20 & 47.35 \\
    \midrule
    \textbf{GPTQ}         & 4-16-16  & 52.78 & 32.85 & 45.10 & 61.71 & 65.18 & 55.56 & 41.15 & 31.00 & 49.67 & 44.89 & 33.86 & 53.05 & 26.23 & 18.90 & 43.71 \\
    \textbf{GPTQ}         & 3-16-16  & 36.91 & 25.60 & 38.66 & 60.18 & 60.66 & 53.67 & 38.54 & 28.80 & 44.84 & 43.48 & 26.45 & 24.95 & 0.61 & 0.00 & 34.53 \\
    \textbf{GPTQ}         & 2-16-16  & 24.87 & 25.26 & 26.43 & 42.63 & 51.41 & 53.43 & 33.83 & 27.00 & 54.33 & 47.60 & 24.72 & 8.50 & 0.00 & 0.00 & 30.00 \\
    \midrule
    \textbf{OmniQuant}    & 4-16-16  & 47.01 & 31.06 & 44.43 & 58.32 & 65.18 & 57.06 & 38.28 & 31.80 & 44.39 & 42.32 & 40.55 & 12.01 & 0.00 & 0.00 & 36.60 \\
    \textbf{OmniQuant}    & 3-16-16  & 44.61 & 26.02 & 38.94 & 63.36 & 61.81 & 54.62 & 36.90 & 29.40 & 43.25 & 43.81 & 30.56 & 10.72 & 0.00 & 0.00 & 34.57 \\
    \textbf{OmniQuant}    & 2-16-16  & 32.41 & 22.27 & 28.55 & 38.62 & 55.22 & 50.51 & 34.54 & 24.60 & 56.73 & 51.28 & 22.92 & 12.20 & 0.00 & 0.00 & 30.70 \\
\bottomrule
  \end{tabular}%
  }
\end{table}

\begin{table}[!ht]
  \centering
  \caption{Detailed performance of weight-activation quantization methods on Qwen3-0.6B.}
  \label{tab:detailed-wakv-quant-on-qwen3-0.6b}
  \setlength{\tabcolsep}{2pt}
  \resizebox{\textwidth}{!}{%
  \begin{tabular}{lccccccccccccccccc}
    \toprule
    \textbf{Methods} & \textbf{W-A-KV} & \textbf{ARC-e} & \textbf{ARC-c} & \textbf{HellaS.} & \textbf{BoolQ} & \textbf{PIQA} & \textbf{WinoG.} & \textbf{SIQA} & \textbf{OBQA} & \textbf{Ethics} & \textbf{Tr.QA2} & \textbf{MMLU} & \textbf{IFEval} & \textbf{GSM8K} & \textbf{HumanE.} & \textbf{Average} \\
    \midrule
    \textbf{BF16}       & 16-16-16 & 56.02 & 34.04 & 47.23 & 64.04 & 67.36 & 56.04 & 39.20 & 31.20 & 47.70 & 42.84 & 40.12 & 58.41 & 41.54 & 37.20 & 47.35 \\
    \midrule
    \textbf{OmniQuant}  & 4-8-8    & 48.11 & 30.46 & 44.06 & 66.24 & 65.07 & 55.88 & 37.82 & 32.00 & 48.90 & 42.07 & 39.11 & 12.01 & 0.00 & 0.00 & 37.27 \\
    \textbf{OmniQuant}  & 3-8-8    & 42.42 & 28.07 & 38.73 & 64.19 & 61.59 & 54.14 & 37.15 & 28.80 & 43.81 & 43.74 & 30.40 & 11.09 & 0.00 & 0.00 & 34.58 \\
    \textbf{OmniQuant}  & 2-8-8    & 32.20 & 21.76 & 27.66 & 38.13 & 54.57 & 50.28 & 33.37 & 26.20 & 56.65 & 51.43 & 22.99 & 11.65 & 0.00 & 0.00 & \underline{30.49} \\
    \midrule
    \textbf{LQER}       & 4-8-8   & 55.64 & 31.83 & 45.21 & 63.76 & 66.05 & 53.51 & 38.43 & 29.80 & 47.86 & 41.85 & 41.13 & 54.16 & 31.61 & 33.54 & 45.31 \\
    \textbf{LQER}       & 3-8-8   & 41.33 & 26.88 & 39.91 & 62.05 & 61.32 & 52.01 & 38.18 & 27.80 & 43.52 & 43.05 & 26.56 & 38.63 & 4.32 & 4.88 & 36.46 \\
    \textbf{LQER}       & 2-8-8   & 27.57 & 25.51 & 26.74 & 53.52 & 53.48 & 50.36 & 33.37 & 27.00 & 44.22 & 49.98 & 23.57 & 11.09 & 0.00 & 0.00 & 30.46 \\
\bottomrule
  \end{tabular}%
  }
\end{table}

\begin{table}[!ht]
  \centering
  \caption{Detailed performance of weight-only quantization methods on Qwen3-1.7B.}
  \label{tab:detailed-weight-only-quant-on-qwen3-1.7b}
  \setlength{\tabcolsep}{2pt}
  \resizebox{\textwidth}{!}{%
  \begin{tabular}{lccccccccccccccccc}
    \toprule
    \textbf{Methods} & \textbf{W-A-KV} & \textbf{ARC-e} & \textbf{ARC-c} & \textbf{HellaS.} & \textbf{BoolQ} & \textbf{PIQA} & \textbf{WinoG.} & \textbf{SIQA} & \textbf{OBQA} & \textbf{Ethics} & \textbf{Tr.QA2} & \textbf{MMLU} & \textbf{IFEval} & \textbf{GSM8K} & \textbf{HumanE.} & \textbf{Average} \\
    \midrule
    \textbf{BF16}         & 16-16-16 & 69.87 & 42.83 & 60.40 & 77.77 & 72.58 & 60.85 & 45.19 & 37.40 & 49.63 & 45.97 & 55.49 & 67.10 & 68.76 & 67.07 & 58.64 \\
    \midrule
    \textbf{GPTQ}         & 4-16-16  & 62.21 & 38.40 & 58.35 & 76.51 & 70.35 & 58.72 & 42.78 & 34.80 & 55.24 & 45.79 & 51.37 & 59.52 & 59.59 & 55.49 & 54.94 \\
    \textbf{GPTQ}         & 3-16-16  & 56.69 & 35.15 & 53.71 & 69.36 & 67.08 & 58.48 & 41.56 & 34.80 & 51.26 & 47.46 & 42.03 & 33.09 & 9.63 & 3.66 & 43.14 \\
    \textbf{GPTQ}         & 2-16-16  & 25.76 & 24.91 & 26.17 & 48.99 & 50.27 & 49.80 & 33.11 & 27.80 & 51.45 & 47.91 & 23.54 & 7.76 & 0.00 & 0.00 & 29.82 \\
    \midrule
    \textbf{OmniQuant}    & 4-16-16  & 69.11 & 41.13 & 58.02 & 79.79 & 71.00 & 62.35 & 44.63 & 36.00 & 52.10 & 44.84 & 52.34 & 15.34 & 0.00 & 0.00 & 44.76 \\
    \textbf{OmniQuant}    & 3-16-16  & 60.61 & 36.01 & 52.49 & 67.00 & 68.55 & 58.33 & 40.84 & 32.80 & 43.64 & 45.24 & 48.95 & 14.60 & 0.00 & 0.00 & 40.65 \\
    \textbf{OmniQuant}    & 2-16-16  & 40.95 & 24.32 & 32.85 & 59.60 & 59.19 & 52.17 & 35.82 & 27.40 & 44.65 & 44.53 & 22.94 & 12.20 & 0.00 & 0.00 & 32.62 \\
\bottomrule
  \end{tabular}%
  }
\end{table}

\begin{table}[!ht]
  \centering
  \caption{Detailed performance of weight-activation quantization methods on Qwen3-1.7B.}
  \label{tab:detailed-weight-activation-quant-on-qwen3-1.7b}
  \setlength{\tabcolsep}{2pt}
  \resizebox{\textwidth}{!}{%
  \begin{tabular}{lccccccccccccccccc}
    \toprule
    \textbf{Methods} & \textbf{W-A-KV} & \textbf{ARC-e} & \textbf{ARC-c} & \textbf{HellaS.} & \textbf{BoolQ} & \textbf{PIQA} & \textbf{WinoG.} & \textbf{SIQA} & \textbf{OBQA} & \textbf{Ethics} & \textbf{Tr.QA2} & \textbf{MMLU} & \textbf{IFEval} & \textbf{GSM8K} & \textbf{HumanE.} & \textbf{Average} \\
    \midrule
    \textbf{BF16}         & 16-16-16 & 69.87 & 42.83 & 60.40 & 77.77 & 72.58 & 60.85 & 45.19 & 37.40 & 49.63 & 45.97 & 55.49 & 67.10 & 68.92 & 67.07 & 58.65 \\
    \midrule
    \textbf{OmniQuant}    & 4-8-8    & 67.42 & 40.36 & 58.46 & 76.54 & 70.57 & 59.83 & 44.42 & 36.80 & 47.09 & 44.20 & 52.96 & 13.12 & 0.00 & 0.00 & 43.70 \\
    \textbf{OmniQuant}    & 3-8-8    & 62.21 & 35.15 & 52.17 & 67.74 & 68.44 & 56.75 & 41.71 & 33.60 & 44.65 & 44.17 & 48.68 & 14.60 & 0.00 & 0.00 & 40.71 \\
    \textbf{OmniQuant}    & 2-8-8    & 39.14 & 22.27 & 32.91 & 62.39 & 58.05 & 51.22 & 35.62 & 28.20 & 43.27 & 48.22 & 22.93 & 11.65 & 0.00 & 0.00 & \underline{32.56} \\
    \midrule
    \textbf{LQER}         & 4-8-8   & 66.75 & 41.38 & 59.88 & 71.47 & 71.60 & 58.56 & 42.63 & 36.40 & 45.28 & 45.79 & 51.67 & 61.92 & 59.67 & 60.98 & 55.28 \\
    \textbf{LQER}         & 3-8-8   & 56.82 & 33.45 & 52.95 & 71.62 & 67.52 & 56.20 & 40.74 & 35.00 & 57.30 & 45.64 & 40.64 & 43.81 & 22.74 & 30.49 & 46.78 \\
    \textbf{LQER}         & 2-8-8   & 27.65 & 24.66 & 26.56 & 59.33 & 50.54 & 49.57 & 34.19 & 25.80 & 45.76 & 49.68 & 22.93 & 12.20 & 0.00 & 0.00 & 30.63 \\
\bottomrule
  \end{tabular}%
  }
\end{table}

\clearpage
\newpage

\subsection{Ablation studies}
\label{apx:detailed-ablation} % Due to space limitations

This section reports the per-task ablation results for Qwen3-0.6B. We evaluate the proposed modules under three distinct bit-width settings, as presented in Tables~\ref{tab:ablation-1}, \ref{tab:ablation-2}, and \ref{tab:ablation-3}. 
We also provide the exact bit-width allocations for the three configurations. For the 2.79-bit and 1.88-bit models, mixed-precision quantization is applied exclusively to the decoder layers, while the embedding and \texttt{lm\_head} layers are maintained at 4-bit. In contrast, the 2.19-bit model applies mixed-precision quantization across all of these components.
Regarding the configuration abbreviations, SG and ST denote super-group and stacked allocations, respectively. The fixed feature distillation baseline aligns the first, second, and last features from the teacher model.

\begin{table}[!ht]
  \centering
  \caption{Detailed performance of quantization methods based on \textsc{EdgeRazor}. The decoder layers are 2.79-bit (50\% 4-bit and 50\% 1.58-bit), while the embedding and \texttt{lm\_head} layers are 4-bit.}
  \label{tab:ablation-1}
  \setlength{\tabcolsep}{2pt}
  \resizebox{\textwidth}{!}{%
\begin{tabular}{lccccccccccccccccc}
    \toprule
    \textbf{Methods} & \textbf{W-A-KV} & \textbf{ARC-e} & \textbf{ARC-c} & \textbf{HellaS.} & \textbf{BoolQ} & \textbf{PIQA} & \textbf{WinoG.} & \textbf{SIQA} & \textbf{OBQA} & \textbf{Ethics} & \textbf{Tr.QA2} & \textbf{MMLU} & \textbf{IFEval} & \textbf{GSM8K} & \textbf{HumanE.} & \textbf{Average} \\
    \midrule
    \rowcolor{gray!20}
    \textbf{$\textsc{EdgeRazor}_\text{SG+A+E}$} & 2.79-16-16 & 51.77 & 28.33 & 37.47 & 70.70 & 63.71 & 54.06 & 40.33 & 28.20 & 55.08 & 42.72 & 36.85 & 51.39 & 26.69 & 31.10 & \textbf{44.17} \\
    \rowcolor{gray!20}
    \textbf{$\textsc{EdgeRazor}_\text{SG+A+E}$} & 2.79-8-8   & 52.10 & 28.50 & 37.36 & 70.58 & 63.93 & 53.12 & 40.12 & 28.60 & 54.97 & 42.82 & 36.44 & 49.54 & 26.99 & 32.32 & \textbf{44.10} \\
    \midrule
    \textbf{$\textsc{EdgeRazor}_\text{ST+A+E}$} & 2.79-16-16 & 55.51 & 30.03 & 37.51 & 54.25 & 64.04 & 53.91 & 40.84 & 28.40 & 58.84 & 44.02 & 37.05 & 46.77 & 28.20 & 26.22 & 43.26 \\
    \textbf{$\textsc{EdgeRazor}_\text{ST+A+E}$} & 2.79-8-8   & 55.47 & 30.29 & 37.56 & 54.04 & 64.09 & 54.30 & 40.79 & 28.40 & 58.77 & 43.93 & 37.05 & 46.03 & 27.45 & 25.00 & 43.08 \\
    \bottomrule
  \end{tabular}%
  }
\end{table}

\begin{table}[!ht]
  \centering
  \caption{Detailed performance of quantization methods based on \textsc{EdgeRazor}. The quantized layers, including decoder, embedding, and \texttt{lm\_head}, are all 2.19-bit (25\% 4-bit and 75\% 1.58-bit).}
  \label{tab:ablation-2}
  \setlength{\tabcolsep}{2pt}
  \resizebox{\textwidth}{!}{%
  \begin{tabular}{lccccccccccccccccc}
    \toprule
    \textbf{Methods} & \textbf{W-A-KV} & \textbf{ARC-e} & \textbf{ARC-c} & \textbf{HellaS.} & \textbf{BoolQ} & \textbf{PIQA} & \textbf{WinoG.} & \textbf{SIQA} & \textbf{OBQA} & \textbf{Ethics} & \textbf{Tr.QA2} & \textbf{MMLU} & \textbf{IFEval} & \textbf{GSM8K} & \textbf{HumanE.} & \textbf{Average} \\
    \midrule
    \rowcolor{gray!20}
    \textbf{$\textsc{EdgeRazor}_\text{SG+A+E}$} & 2.19-16-16 & 50.17 & 29.44 & 34.21 & 63.88 & 62.89 & 50.83 & 37.00 & 29.80 & 48.45 & 43.59 & 31.77 & 38.82 & 24.64 & 24.39 & \textbf{40.71} \\
    \rowcolor{gray!20}
    \textbf{$\textsc{EdgeRazor}_\text{SG+A+E}$} & 2.19-8-8   & 49.24 & 28.67 & 34.28 & 63.98 & 61.86 & 50.43 & 36.69 & 29.80 & 47.12 & 44.14 & 31.71 & 39.19 & 22.90 & 21.95 & \textbf{40.14} \\
    \midrule
    \textbf{$\textsc{EdgeRazor}_\text{SG+A+C}$} & 2.19-16-16 & 52.27 & 27.82 & 33.59 & 65.05 & 62.46 & 50.20 & 37.92 & 28.00 & 43.29 & 44.54 & 27.05 & 40.30 & 20.39 & 21.34 & 39.59 \\
    \textbf{$\textsc{EdgeRazor}_\text{SG+A+C}$} & 2.19-8-8   & 52.31 & 27.47 & 33.70 & 64.65 & 62.19 & 50.51 & 37.77 & 27.80 & 43.29 & 44.59 & 27.13 & 40.30 & 20.24 & 22.56 & 39.61 \\
    \midrule
    \textbf{$\textsc{EdgeRazor}_\text{SG+F+E}$} & 2.19-16-16 & 49.03 & 27.05 & 34.33 & 59.57 & 62.51 & 51.46 & 38.28 & 30.20 & 53.63 & 45.34 & 27.88 & 36.97 & 27.37 & 20.12 & 40.27 \\
    \textbf{$\textsc{EdgeRazor}_\text{SG+F+E}$} & 2.19-8-8   & 47.77 & 26.96 & 34.02 & 59.33 & 61.86 & 52.09 & 38.13 & 30.60 & 53.22 & 44.98 & 27.89 & 35.67 & 27.60 & 18.90 & 39.93 \\
    \midrule
    \textbf{$\textsc{EdgeRazor}_\text{SG+F+F}$} & 2.19-16-16 & 49.15 & 26.11 & 30.04 & 52.29 & 63.38 & 51.70 & 38.23 & 29.00 & 55.86 & 45.85 & 29.13 & 36.04 & 22.59 & 20.12 & 39.25 \\
    \textbf{$\textsc{EdgeRazor}_\text{SG+F+F}$} & 2.19-8-8   & 48.99 & 26.11 & 33.24 & 51.83 & 62.89 & 51.38 & 38.13 & 28.40 & 55.77 & 45.92 & 28.98 & 37.15 & 21.15 & 23.17 & 39.51 \\
    \bottomrule
  \end{tabular}%
  }
\end{table}

\begin{table}[!ht]
  \centering
  \caption{Detailed performance of quantization methods based on \textsc{EdgeRazor}. The decoder layers are 1.88-bit (12.5\% 4-bit and 87.5\% 1.58-bit), while the embedding and \texttt{lm\_head} layers are 4-bit.}
  \label{tab:ablation-3}
  \setlength{\tabcolsep}{2pt}
  \resizebox{\textwidth}{!}{%
  \begin{tabular}{lccccccccccccccccc}
    \toprule
    \textbf{Methods} & \textbf{W-A-KV} & \textbf{ARC-e} & \textbf{ARC-c} & \textbf{HellaS.} & \textbf{BoolQ} & \textbf{PIQA} & \textbf{WinoG.} & \textbf{SIQA} & \textbf{OBQA} & \textbf{Ethics} & \textbf{Tr.QA2} & \textbf{MMLU} & \textbf{IFEval} & \textbf{GSM8K} & \textbf{HumanE.} & \textbf{Average} \\
    \midrule
    \rowcolor{gray!20}
    \textbf{$\textsc{EdgeRazor}_\text{SG+A+E}$} & 1.88-16-16 & 51.22 & 27.73 & 34.21 & 66.91 & 63.66 & 53.35 & 38.43 & 27.60 & 55.92 & 43.80 & 28.78 & 42.51 & 25.09 & 23.17 & \textbf{41.60} \\
    \rowcolor{gray!20}
    \textbf{$\textsc{EdgeRazor}_\text{SG+A+E}$} & 1.88-8-8   & 51.47 & 27.99 & 34.22 & 66.85 & 63.49 & 53.04 & 38.02 & 27.40 & 55.92 & 43.88 & 29.56 & 44.55 & 25.09 & 23.17 & \textbf{41.76} \\
    \midrule
    \textbf{$\textsc{EdgeRazor}_\text{SG+A+C}$} & 1.88-16-16 & 49.20 & 27.56 & 34.64 & 65.05 & 61.75 & 53.67 & 39.36 & 30.00 & 54.72 & 44.35 & 30.90 & 39.56 & 23.05 & 19.51 & 40.95 \\
    \textbf{$\textsc{EdgeRazor}_\text{SG+A+C}$} & 1.88-8-8   & 49.16 & 27.47 & 34.50 & 64.95 & 61.75 & 54.06 & 39.20 & 29.40 & 54.80 & 44.47 & 30.59 & 37.34 & 22.21 & 21.95 & 40.85 \\
    \midrule
    \textbf{$\textsc{EdgeRazor}_\text{SG+F+E}$} & 1.88-16-16 & 48.57 & 26.45 & 33.60 & 58.65 & 61.53 & 53.20 & 39.30 & 29.40 & 56.46 & 43.67 & 33.21 & 40.67 & 20.77 & 20.12 & 40.40 \\
    \textbf{$\textsc{EdgeRazor}_\text{SG+F+E}$} & 1.88-8-8   & 48.15 & 25.94 & 33.75 & 58.50 & 60.83 & 52.72 & 38.54 & 29.40 & 56.44 & 43.50 & 33.33 & 41.04 & 19.86 & 21.34 & 40.24 \\
    \midrule
    \textbf{$\textsc{EdgeRazor}_\text{SG+F+F}$} & 1.88-16-16 & 47.90 & 27.39 & 34.23 & 61.83 & 61.43 & 52.49 & 38.23 & 26.80 & 54.12 & 47.87 & 26.46 & 39.37 & 20.55 & 18.90 & 39.83 \\
    \textbf{$\textsc{EdgeRazor}_\text{SG+F+F}$} & 1.88-8-8   & 48.06 & 26.96 & 34.30 & 61.77 & 61.37 & 51.93 & 38.38 & 26.60 & 52.13 & 47.53 & 25.99 & 39.74 & 20.92 & 20.12 & 39.70 \\
    \bottomrule
  \end{tabular}%
  }
\end{table}

% ---------------------------------------------------------------------------------------

\subsection{Efficiency}
\label{apx:efficiency}

\begin{table}[!t]
  \centering
  \caption{Compression comparison of quantization methods on MobileLLM-350M.}
  \label{tab:efficiency_compr_mobilellm}
  \setlength{\tabcolsep}{3pt}
  \begin{tabular}{lcccccccc}
    \toprule
    \multirow{2.5}{*}{\textbf{Methods}} & \multirow{2.5}{*}{\textbf{Bit-widths}} & \multirow{2.5}{*}{\textbf{\shortstack{Group \\ sizes}}} & \multicolumn{3}{c}{\textbf{Quantized layers}} & \multirow{2.5}{*}{\textbf{\shortstack{Quantization \\ proportions ($\uparrow$)}}} & \multirow{2.5}{*}{\textbf{\shortstack{Compression \\ ratios ($\uparrow$)}}} \\
    \cmidrule(lr){4-6}
    & & & \textbf{decoder} & \textbf{emb} & \textbf{lm\_head} & & \\
    \midrule
    \textbf{BF16} & 16   & --  & --         & --         & --         & 00.00\%               & 1.00\texttimes \\
    \midrule
    \rowcolor{gray!20}
    & 4    & 256 & \checkmark & \checkmark & \checkmark & \textbf{99.99\%} & \textbf{3.76\texttimes} \\
    \rowcolor{gray!20}
    & 2.79 & 256 & \checkmark & \checkmark & \checkmark & \textbf{99.99\%} & \textbf{4.94\texttimes} \\
    \rowcolor{gray!20}
    & 1.88 & 256 & \checkmark & \checkmark & \checkmark & \textbf{99.99\%} & \textbf{6.46\texttimes} \\
    \rowcolor{gray!20}
    \multirow{-4}{*}{\textbf{\textsc{EdgeRazor}}}
        & 1.58 & 256 & \checkmark & \checkmark & \checkmark & \textbf{99.99\%} & \textbf{7.19\texttimes} \\
    \midrule
    \multirow{4}{*}{\textbf{Other methods}}
        & 4    & 128 & \checkmark & \texttimes & \texttimes & 83.66\% & 2.64\texttimes \\
\bottomrule
  \end{tabular}
  \vspace{-10pt}
\end{table}

\begin{table}[!t]
  \centering
  \caption{Compression comparison of quantization methods on Qwen3-0.6B.}
  \label{tab:efficiency_compr_qwen3-0.6b}
  \setlength{\tabcolsep}{3pt}
  \begin{tabular}{lcccccccc}
    \toprule
    \multirow{2.5}{*}{\textbf{Methods}} & \multirow{2.5}{*}{\textbf{Bit-widths}} & \multirow{2.5}{*}{\textbf{\shortstack{Group \\ sizes}}} & \multicolumn{3}{c}{\textbf{Quantized layers}} & \multirow{2.5}{*}{\textbf{\shortstack{Quantization \\ proportions ($\uparrow$)}}} & \multirow{2.5}{*}{\textbf{\shortstack{Compression \\ ratios ($\uparrow$)}}} \\
    \cmidrule(lr){4-6}
    & & & \textbf{decoder} & \textbf{emb} & \textbf{{lm\_head}} & & \\
    \midrule
    \textbf{BF16} & 16   & --  & --         & --         & --         & 00.00\%               & 1.00\texttimes \\
    \midrule
    \rowcolor{gray!20}
    & 4    & 256 & \checkmark & \checkmark & \checkmark & \textbf{99.99\%} & \textbf{3.94\texttimes} \\
    \rowcolor{gray!20}
    & 2.79 & 256 & \checkmark & \checkmark & \checkmark & \textbf{99.99\%} & \textbf{5.05\texttimes} \\
    \rowcolor{gray!20}
    & 1.88 & 256 & \checkmark & \checkmark & \checkmark & \textbf{99.99\%} & \textbf{6.41\texttimes} \\
    \rowcolor{gray!20}
    \multirow{-4}{*}{\textbf{\textsc{EdgeRazor}}}
        & 1.58 & 256 & \checkmark & \checkmark & \checkmark & \textbf{99.99\%} & \textbf{7.04\texttimes} \\
    \midrule
    \multirow{4}{*}{\textbf{Other methods}}
        & 4    & 128 & \checkmark & \texttimes & \texttimes & 73.89\% & 2.21\texttimes \\
\bottomrule
  \end{tabular}
  \vspace{-10pt}
\end{table}

\begin{table}[!t]
  \centering
  \caption{Compression comparison of quantization methods on Qwen3-1.7B.}
  \label{tab:efficiency_compr_qwne3-1.7b}
  \setlength{\tabcolsep}{3pt}
  \begin{tabular}{lcccccccc}
    \toprule
    \multirow{2.5}{*}{\textbf{Methods}} & \multirow{2.5}{*}{\textbf{Bit-widths}} & \multirow{2.5}{*}{\textbf{\shortstack{Group \\ sizes}}} & \multicolumn{3}{c}{\textbf{Quantized layers}} & \multirow{2.5}{*}{\textbf{\shortstack{Quantization \\ proportions ($\uparrow$)}}} & \multirow{2.5}{*}{\textbf{\shortstack{Compression \\ ratios ($\uparrow$)}}} \\
    \cmidrule(lr){4-6}
    & & & \textbf{decoder} & \textbf{emb} & \textbf{lm\_head} & & \\
    \midrule
    \textbf{BF16} & 16   & --  & --         & --         & --         & 00.00\%               & 1.00\texttimes \\
    \midrule
    \rowcolor{gray!20}
    & 4    & 256 & \checkmark & \checkmark & \checkmark & \textbf{99.99\%} & \textbf{3.94\texttimes} \\
    \rowcolor{gray!20}
    & 2.79 & 256 & \checkmark & \checkmark & \checkmark & \textbf{99.99\%} & \textbf{5.21\texttimes} \\
    \rowcolor{gray!20}
    & 1.88 & 256 & \checkmark & \checkmark & \checkmark & \textbf{99.99\%} & \textbf{6.88\texttimes} \\
    \rowcolor{gray!20}
    \multirow{-4}{*}{\textbf{\textsc{EdgeRazor}}}
        & 1.58 & 256 & \checkmark & \checkmark & \checkmark & \textbf{99.99\%} & \textbf{7.69\texttimes} \\
    \midrule
    \multirow{4}{*}{\textbf{Other methods}}
        & 4    & 128 & \checkmark & \texttimes & \texttimes & 81.91\% & 2.55\texttimes \\
\bottomrule
  \end{tabular}
\end{table}

\begin{table}[!t]
  \centering
  \caption{Compression comparison of quantization methods on Qwen2.5-Omni-7B.}
  \label{tab:efficiency_compr_qwne2.5}
  \setlength{\tabcolsep}{3pt}
  \begin{tabular}{lcccccccc}
    \toprule
    \multirow{2.5}{*}{\textbf{Methods}} & \multirow{2.5}{*}{\textbf{Bit-widths}} & \multirow{2.5}{*}{\textbf{\shortstack{Group \\ sizes}}} & \multicolumn{4}{c}{\textbf{Quantized layers}} & \multirow{2.5}{*}{\textbf{\shortstack{Prop. ($\uparrow$)}}} & \multirow{2.5}{*}{\textbf{\shortstack{Compr. ($\uparrow$)}}} \\
    \cmidrule(lr){4-7}
    & & & \textbf{vision} & \textbf{decoder} & \textbf{emb} & \textbf{lm\_head} & & \\
    \midrule
    \textbf{BF16} & 16   & --  & --         & --         & --         & --         & 00.00\%               & 1.00\texttimes \\
    \midrule
    \rowcolor{gray!20}
    \textbf{\textsc{EdgeRazor}} & 4    & 64 & \checkmark & \checkmark & \checkmark & \checkmark & \textbf{96.73\%} & \textbf{3.45\texttimes} \\
    \textbf{GPTQ and AWQ} & 4    & 128 & \texttimes & \checkmark & \texttimes & \texttimes & 79.81\% & 2.45\texttimes \\
    \bottomrule
  \end{tabular}
\end{table}

\begin{table}[!ht]
\caption{Inference comparison of text LLMs on two chips: Apple M4 Pro and Intel i9-14900K.}
\label{tab:efficiency_inference_detailed}
\centering
\small
\setlength{\tabcolsep}{4pt}
\begin{tabular}{lllcccccc}
\toprule
\textbf{Chips} & \textbf{Models} & \textbf{Methods} & \makecell{\textbf{Weight} \\ \textbf{types}} & \makecell{\textbf{KV} \\ \textbf{types}} & \makecell{\textbf{Storage} \\ \textbf{(GB)}} & \makecell{\textbf{Memory} \\ \textbf{(GB)}} & \makecell{\textbf{Prefilling} \\ \textbf{(tok/s)}} & \makecell{\textbf{Decoding} \\ \textbf{(tok/s)}} \\
\midrule

% --- Apple M4 Pro Section ---
\multirow{18}{*}{\rotatebox{90}{\textbf{Apple M4 Pro}}} 
& \multirow{6}{*}{\textbf{MobileLLM-350M}} 
& \textbf{BF16} & BF16 & BF16 & 0.70 & 0.79 & 614.23 & 37.99 \\
& & \textbf{Q4\_K} & Q4\_K & Q8\_0 & 0.25 & 0.33 & 1019.88 & 325.81 \\
& & \cellcolor{gray!20}\textbf{\textsc{EdgeRazor}} & \cellcolor{gray!20}Q4\_0 & \cellcolor{gray!20}Q8\_0 & \cellcolor{gray!20}0.21 & \cellcolor{gray!20}0.29 & \cellcolor{gray!20}\textbf{1901.14} & \cellcolor{gray!20}426.47 \\
& & \textbf{Q2\_K} & Q2\_K & Q8\_0 & 0.20 & 0.28 & 1505.05 & 398.31 \\
& & \cellcolor{gray!20}\textbf{\textsc{EdgeRazor}} & \cellcolor{gray!20}TQ1\_0 & \cellcolor{gray!20}Q8\_0 & \cellcolor{gray!20}\textbf{0.17} & \cellcolor{gray!20}\textbf{0.23} & \cellcolor{gray!20}1583.24 & \cellcolor{gray!20}439.52 \\
& & \cellcolor{gray!20}\textbf{\textsc{EdgeRazor}} & \cellcolor{gray!20}TQ2\_0 & \cellcolor{gray!20}Q8\_0 & \cellcolor{gray!20}0.18 & \cellcolor{gray!20}0.24 & \cellcolor{gray!20}1604.57 & \cellcolor{gray!20}\textbf{458.66} \\
\cmidrule{2-9}
& \multirow{6}{*}{\textbf{Qwen3-0.6B}} 
& \textbf{BF16} & BF16 & BF16 & 1.11 & 1.46 & 337.99 & 20.91 \\
\bottomrule
\end{tabular}
\end{table}

In this section, we present all efficiency metrics for MobileLLM-350M, Qwen3-0.6B, Qwen3-1.7B, and Qwen2.5-Omni-7B on the Apple M4 Pro and Intel i9-14900K chips. Since the multimodal projector feature is experimental in the inference framework \texttt{llama.cpp} and is not compatible with the \texttt{llama-bench} tool, Qwen2.5-Omni-7B is not included in the inference benchmark. For deployment benchmarks on text LLMs, we use the default thread count, a prompt length of 512, a generation length of 512, a batch size of 4096, and 100 repetitions.
Tables~\ref{tab:efficiency_compr_mobilellm}, \ref{tab:efficiency_compr_qwen3-0.6b}, \ref{tab:efficiency_compr_qwne3-1.7b}, and \ref{tab:efficiency_compr_qwne2.5} show that \textsc{EdgeRazor} attains the highest quantization proportions and compression ratios across diverse LLMs. Table~\ref{tab:efficiency_inference_detailed} further indicates that \textsc{EdgeRazor} achieves promising efficiency with compatible \texttt{llama.cpp} precision types on two chips.

% ---------------------------------------------------------------------------------------

%%%%%%%%%%%%%%%%%%%%%%%%%%%%%%%%%%%%%%%%%%%%%%%%%%%%%%%%%%%%

\clearpage
\newpage

\section{Generalization on held-out benchmarks}
\label{apx:generalization-held-out}

\begin{table}[!ht]
  \centering
  \caption{Average performance of weight-only quantization methods on Qwen3 models.}
  \label{tab:qwen3-heldout-benchmark-w}
  \small
  \begin{tabular}{lccccc}
    \toprule
    \textbf{Methods} & \textbf{W} & \textbf{A} & \textbf{KV} & \textbf{Qwen3-0.6B} & \textbf{Qwen3-1.7B} \\
    \midrule
    \textbf{BF16}        & 16 & 16 & 16 & 44.32 & 64.61 \\
    \midrule
    \textbf{GPTQ}        & 4  & 16 & 16 & 33.01 & 56.49 \\
    \textbf{OmniQuant}   & 4  & 16 & 16 & 13.14 & 16.92 \\
    \textbf{AWQ}         & 4  & 16 & 16 & 40.11 & 60.96 \\
    \textbf{AQLM}        & 4  & 16 & 16 & \textbf{42.41} & \textbf{62.69} \\
    \textbf{QuIP\#}      & 4  & 16 & 16 & 8.81  & 9.46 \\
    \textbf{AutoRound}   & 4  & 16 & 16 & 41.71 & 61.10 \\
    \textbf{VPTQ}        & 4  & 16 & 16 & 28.63 & \underline{61.42} \\
    \textbf{GPTAQ}       & 4  & 16 & 16 & 35.12 & 59.16 \\
    \textbf{Q-Palette}   & 4  & 16 & 16 & 26.32 & 38.05 \\
    \textbf{EfficientQAT}& 4  & 16 & 16 & 30.70 & 44.21 \\
    \rowcolor{gray!20}
    \textbf{\textsc{EdgeRazor}}   & 4  & 16 & 16 & \underline{42.24} & 59.84 \\
    \midrule
    \textbf{GPTQ}        & 3  & 16 & 16 & 13.00 & 22.10 \\
    \textbf{OmniQuant}   & 3  & 16 & 16 & 10.32 & 15.89 \\
    \textbf{AWQ}         & 3  & 16 & 16 & 14.40 & 39.44 \\
    \textbf{AQLM}        & 3  & 16 & 16 & 23.12 & 49.05 \\
    \textbf{QuIP\#}      & 3  & 16 & 16 & 12.45 & 11.45 \\
    \textbf{AutoRound}   & 3  & 16 & 16 & \underline{28.23} & \underline{49.01} \\
    \textbf{VPTQ}        & 3  & 16 & 16 & 18.73 & 35.94 \\
    \textbf{GPTAQ}       & 3  & 16 & 16 & 14.15 & 28.98 \\
    \textbf{Slim-LLM+}   & 3  & 16 & 16 & 9.22  & 37.37 \\
    \textbf{Q-Palette}   & 3.25 & 16 & 16 & 19.82 & 33.47 \\
    \textbf{EfficientQAT}& 3  & 16 & 16 & 22.45 & 39.69 \\
    \rowcolor{gray!20}
    \textbf{\textsc{EdgeRazor}}   & 2.79 & 16 & 16 & \textbf{36.51} & \textbf{52.23} \\
    \midrule
    \textbf{GPTQ}        & 2  & 16 & 16 & 8.31  & 7.83 \\
    \textbf{OmniQuant}   & 2  & 16 & 16 & 8.78  & 8.79 \\
    \textbf{AWQ}         & 2  & 16 & 16 & 9.45  & 9.26 \\
    \textbf{AQLM}        & 2  & 16 & 16 & \underline{19.26} & 21.92 \\
    \textbf{QuIP\#}      & 2  & 16 & 16 & 8.33  & 9.48 \\
    \textbf{AutoRound}   & 2  & 16 & 16 & 7.72  & 10.87 \\
    \textbf{VPTQ}        & 2  & 16 & 16 & 8.14  & 8.82 \\
    \textbf{QTIP}        & 2  & 16 & 16 & 12.48 & \underline{34.36} \\
    \textbf{GPTAQ}       & 2  & 16 & 16 & 8.01  & 7.87 \\
    \textbf{Slim-LLM+}   & 2  & 16 & 16 & 8.52  & 9.39 \\
    \textbf{Q-Palette}   & 2  & 16 & 16 & 8.31  & 10.55 \\
    \textbf{EfficientQAT}& 2  & 16 & 16 & 6.82  & 12.94 \\
    \rowcolor{gray!20}
    \textbf{\textsc{EdgeRazor}}   & 1.88 & 16 & 16 & \textbf{29.89} & \textbf{39.69} \\
    \midrule
    \textbf{BiLLM}       & 1.06 & 16 & 16 & \underline{8.98}  & 8.89 \\
    \textbf{ARB-LLM}     & 1  & 16 & 16 & 9.02  & 6.55 \\
    \textbf{Q-Palette}   & 1.75 & 16 & 16 & 9.02  & \underline{8.91} \\
    \rowcolor{gray!20}
    \textbf{\textsc{EdgeRazor}}   & 1.58 & 16 & 16 & \textbf{26.90} & \textbf{32.69} \\
    \bottomrule
  \end{tabular}
\end{table}

\begin{table}[!ht]
  \centering
  \caption{Average performance of weight-activation quantization methods on Qwen3 models.}
  \label{tab:qwen3-heldout-benchmark-wakv}
  \small
  \begin{tabular}{lccccc}
    \toprule
    \textbf{Methods} & \textbf{W} & \textbf{A} & \textbf{KV} & \textbf{Qwen3-0.6B} & \textbf{Qwen3-1.7B} \\
    \midrule
    \textbf{BF16}      & 16 & 16 & 16 & 44.32 & 64.65 \\
    \midrule
    \textbf{OmniQuant} & 4  & 8  & 8  & 12.78 & 16.52 \\
    \textbf{LQER}      & 4  & 8  & 8  & \underline{40.11} & 58.56 \\
    \textbf{QuaRot}    & 4  & 8  & 8  & 8.24  & 8.67 \\
    \textbf{ABQ-LLM}   & 4  & 8  & 8  & 34.14 & 16.24 \\
    \textbf{SpinQuant} & 4  & 8  & 8  & 27.40 & 59.02 \\
    \textbf{QoQ}       & 4  & 8  & 4  & 8.97  & 9.20 \\
    \textbf{FlatQuant} & 4  & 8  & 8  & 37.26 & \textbf{62.67} \\
    \rowcolor{gray!20}
    \textbf{\textsc{EdgeRazor}} & 4  & 8  & 8  & \textbf{42.48} & \underline{60.56} \\
    \midrule
    \textbf{OmniQuant} & 3  & 8  & 8  & 10.37 & 15.82 \\
    \textbf{LQER}      & 3  & 8  & 8  & 18.60 & 34.42 \\
    \textbf{QuaRot}    & 3  & 8  & 8  & 8.34  & 8.56 \\
    \textbf{ABQ-LLM}   & 3  & 8  & 8  & 8.68  & 13.42 \\
    \textbf{SpinQuant} & 3  & 8  & 8  & 15.06 & 33.95 \\
    \textbf{FlatQuant} & 3  & 8  & 8  & \underline{18.85} & \underline{40.32} \\
    \rowcolor{gray!20}
    \textbf{\textsc{EdgeRazor}} & 2.79 & 8  & 8  & \textbf{36.32} & \textbf{51.23} \\
    \midrule
    \textbf{OmniQuant} & 2  & 8  & 8  & 8.66  & 8.65 \\
    \textbf{LQER}      & 2  & 8  & 8  & 8.67  & 8.78 \\
    \textbf{QuaRot}    & 2  & 8  & 8  & 8.05  & 8.63 \\
    \textbf{ABQ-LLM}   & 2.32 & 8  & 8  & \underline{9.09}  & 8.66 \\
    \textbf{SpinQuant} & 2  & 8  & 8  & 9.03  & 6.77 \\
    \textbf{FlatQuant} & 2  & 8  & 8  & 8.56  & \underline{8.82} \\
    \rowcolor{gray!20}
    \textbf{\textsc{EdgeRazor}} & 1.88 & 8  & 8  & \textbf{30.59} & \textbf{39.84} \\
    \rowcolor{gray!20}
    \textbf{\textsc{EdgeRazor}} & 1.58 & 8  & 8  & \textbf{27.28} & \textbf{32.33} \\
    \bottomrule
  \end{tabular}
\end{table}

The evaluation scores reported in the main text demonstrate the excellent performance of \textsc{EdgeRazor} across extensive downstream tasks. When the training mixture incorporates the training splits of several commonsense reasoning tasks, the main results constitute a comprehensive evaluation that integrates supervised domain adaptations and out-of-domain evaluations. To rigorously validate the generalization capabilities of our proposed framework, this section isolates the evaluation exclusively to held-out benchmarks. We assess knowledge with MMLU, instruction following with IFEval, and code generation with HumanEval in a zero-shot setting, and mathematical reasoning with GSM8K in a five-shot setting. We exclude Qwen2.5-Omni-7B due to its reliance on teacher-distilled data and the base LLM MobileLLM-350M as it inherently lacks the capacity for complex reasoning and coding, yielding uninformative near-zero performance as detailed in Tables~\ref{tab:detailed-weight-only-quant-on-mobilellm350m} and \ref{tab:detailed-wakv-quant-on-mobilellm350m}. Consequently, Tables~\ref{tab:qwen3-heldout-benchmark-w} and \ref{tab:qwen3-heldout-benchmark-wakv} collect the average performance of these challenging held-out benchmarks on the instruction-tuned Qwen3 models. Notably, the best performance is indicated in \textbf{bold}, and the second-best is \underline{underlined}.

There are three observations.
(1) In Tables~\ref{tab:qwen3-heldout-benchmark-w} and \ref{tab:qwen3-heldout-benchmark-wakv}, existing quantization baselines experience severe performance degradation on complex cognitive tasks when compressed below 4-bit, with the strongest 2-bit weight-activation baselines ABQ-LLM and FlatQuant collapsing to scores below 10 points;
(2) \textsc{EdgeRazor} at 2.79-bit establishes a massive performance advantage over all 3-bit baselines. In Table~\ref{tab:qwen3-heldout-benchmark-w}, it outperforms the strongest 3-bit competitor AutoRound by 8.28 and 3.22 points on Qwen3-0.6B and Qwen3-1.7B. Under the weight-activation setup in Table~\ref{tab:qwen3-heldout-benchmark-wakv}, this margin widens, surpassing the leading 3-bit FlatQuant by 17.47 and 10.91 points;
(3) In the sub-2-bit regime where competing baselines fail to generate coherent responses, \textsc{EdgeRazor} robustly preserves foundational cognitive and generative capabilities. In Table~\ref{tab:qwen3-heldout-benchmark-wakv}, \textsc{EdgeRazor} at 1.58-bit achieves 27.28 and 32.33 on Qwen3-0.6B and Qwen3-1.7B, exceeding the best 2-bit baselines ABQ-LLM and FlatQuant by 18.19 and 23.51 points.

These observations demonstrate that the performance advantages of \textsc{EdgeRazor} derive from generalized capability enhancements rather than mere supervised domain adaptations. The results explicitly confirm the robustness and generalization of our proposed framework on completely held-out and complex tasks.

%%%%%%%%%%%%%%%%%%%%%%%%%%%%%%%%%%%%%%%%%%%%%%%%%%%%%%%%%%%%

\clearpage
\newpage
\section{Capability comparison of low-bit LLMs}
\label{apx:capability-low-bit-llms} % comparison: 1.58-bit Qwen3 response

Beyond commonsense question answering, instruction-tuned LLMs are expected to exhibit a broad spectrum of advanced capabilities, among which code generation is one of the most demanding: it requires the model to jointly preserve natural-language understanding, symbolic reasoning, and syntactic precision. As such, code generation serves as a particularly stringent stress test for quantization methods operating in the ultra-low-bit regime.

In this section, we empirically investigate how different quantization schemes affect the coding capability of ultra-low-bit LLMs. To make the comparison informative, we select the best baselines identified in Appendix~\ref{apx:effectiveness}, namely AQLM~(W2-A16-KV16) and OmniQuant~(W2-A8-KV8), \texttt{llama.cpp Q2\_K} as the PTQ baseline from the open-sourced inference framework, and contrast them with our proposed \textsc{EdgeRazor} at an even more aggressive bit-width of W1.58-A8-KV8.
All methods are applied on the same 16-bit Qwen3-0.6B and evaluated with the zero-shot prompt from the HumanEval benchmark.

Table~\ref{tab:coding-cases} presents a representative case study on the \texttt{count\_distinct\_characters} problem.
This generation comparison corroborates the quantitative findings reported in Appendix~\ref{apx:effectiveness}: existing ultra-low-bit quantization methods either strip the model of its instruction-following capability (AQLM) or drive it into incoherent token distributions (OmniQuant), while \textsc{EdgeRazor} substantially preserves high-level coding capability at sub-2-bit weight precision. We attribute this robustness to our framework, \textsc{EdgeRazor}, integrating mixed-precision quantization and distillation schemes to mitigate catastrophic degradation in capabilities.

\begin{table}[!ht]
  \centering
  \caption{Zero-shot coding cases of ultra-low-bit Qwen3-0.6B. The zero-shot prompt is from the HumanEval benchmark.}
  \label{tab:coding-cases}
  \begin{tabularx}{\textwidth}{@{}>{\raggedright\arraybackslash}p{0.18\textwidth} X@{}}
    \toprule
    \textbf{Prompt} &
    Write a solution to the following problem and make sure that it passes the tests:\newline
    \texttt{\`{}\`{}\`{}python}\newline
    \texttt{def count\_distinct\_characters(string: str) -> int:}\newline
    \texttt{\quad""" Given a string, find out how many distinct characters (regardless of case) it consists of}\newline
    \texttt{\quad>{}>{}> count\_distinct\_characters('xyzXYZ')}\newline
    \texttt{\quad 3}\newline
    \texttt{\quad>{}>{}> count\_distinct\_characters('Jerry')}\newline
    \texttt{\quad 4}\newline
    \texttt{\quad"""}\newline
    \texttt{\`{}\`{}\`{}} \\
    % AQLM
    \midrule
    \textbf{AQLM W2-A16-KV16} &
    \textcolor{red}{\textbf{Filtered Generation:}}\newline
    % \texttt{def count\_distinct\_characters(string: str) -> int:}\newline
    \texttt{The solution passes the tests.} \\
    % OmniQuant
    \midrule
    \textbf{OmniQuant W2-A8-KV8} &
    \textcolor{red}{\textbf{Filtered Generation:}}\newline
    % \texttt{def count\_distinct\_characters(string: str) -> int:}\newline
    \texttt{\textbackslash ubc31\textbackslash ub155\textbackslash ub155\textbackslash ub155\textbackslash ub155\textbackslash ub155\textbackslash ub155\textbackslash ub155...} \\
    % Q2_K
    \midrule
    \textbf{{llama.cpp Q2\_K} W2-A8-KV8} &
    \textcolor{red}{\textbf{Filtered Generation:}}\newline
    % \texttt{def count\_distinct\_characters(string: str) -> int:}\newline
    % \texttt{<think>}\newline
    % \texttt{</think>}\newline
    % \newline
    \texttt{\# The code is written correctly and passes the tests. The function implementation is as described.} \\
    % EdgeRazor
    \midrule
    \textbf{\textsc{EdgeRazor} W1.58-A8-KV8} &
    \textcolor{red}{\textbf{Filtered Generation:}}\newline
    % \texttt{def count\_distinct\_characters(string: str) -> int:}\newline
    \texttt{\# Convert the string to lowercase to remove case differences}\newline
    \texttt{cleaned\_string = string.lower()}\newline
    \newline
    \texttt{\# Count the distinct characters}\newline
    \texttt{distinct\_chars = set(cleaned\_string)}\newline
    \texttt{return len(distinct\_chars)} \\
    \bottomrule
  \end{tabularx}
\end{table}

%%%%%%%%%%%%%%%%%%%%%%%%%%%%%%%%%%%%%%%%%%%%%%%%%%%%%%%%%%%%

% \newpage
% \section*{NeurIPS Paper Checklist}
% No "NeurIPS Paper Checklist" for arxiv edition

\end{document}